\documentclass[lettersize,journal]{IEEEtran}
\usepackage{amsmath,amsfonts}
\usepackage{array}
\usepackage[caption=false,font=small,labelfont=sf,textfont=sf]{subfig}
\usepackage{textcomp}
\usepackage{stfloats}
\usepackage{url}
\usepackage{verbatim}
\usepackage{graphicx}
\usepackage{cite}

\usepackage{multirow}
\usepackage{multicol}
\usepackage{color}
\usepackage[noend]{algpseudocode}
\usepackage{algorithmicx,algorithm}

\usepackage{colortbl}
\definecolor{mygray}{gray}{.9}
\usepackage{amssymb}

\begin{document}

\title{Offline-Online Associated Camera-Aware Proxies for Unsupervised Person Re-identification}

\author{Menglin Wang, Jiachen Li, Baisheng Lai, Xiaojin Gong,~\IEEEmembership{Member,~IEEE}, Xian-Sheng Hua,~\IEEEmembership{Fellow,~IEEE}

\thanks{This work was conducted when M. Wang was with Zhejiang University. Email: lynnwang6875@gmail.com.}
\thanks{J. Li and X. Gong are with the College of Information Science and Electronic Engineering, Zhejiang University, and Zhejiang Provincial Key Laboratory of Information Processing, Communication and Networking (IPCAN), Hangzhou 310027, China. Email: lijiachen\_isee@zju.edu.cn, gongxj@zju.edu.cn.}% <-this % stops a space
\thanks{B. Lai and X.-S. Hua are with Alibaba Group, Hangzhou, China. Email: baisheng.lbs@alibaba-inc.com, huaxiansheng@gmail.com.}
\thanks{X. Gong is the corresponding author.}% <-this % stops a space
%\thanks{Manuscript received xxxx, 2022; revised xxxx, 2022.}
}

\markboth{Journal of \LaTeX\ Class Files,~Vol.~14, No.~8, August~2021}%
{Shell \MakeLowercase{\textit{et al.}}: A Sample Article Using IEEEtran.cls for IEEE Journals}


\maketitle

\begin{abstract}
Recently, unsupervised person re-identification (Re-ID) has received increasing research attention due to its potential for label-free applications. A promising way to address unsupervised Re-ID is clustering-based, which generates pseudo labels by clustering and uses the pseudo labels to train a Re-ID model iteratively. However, most clustering-based methods take each cluster as a pseudo identity class, neglecting the intra-cluster variance mainly caused by the change of cameras. To address this issue, we propose to split each single cluster into multiple proxies according to camera views. The camera-aware proxies explicitly capture local structures within clusters, by which the intra-ID variance and inter-ID similarity can be better tackled. Assisted with the camera-aware proxies, we design two proxy-level contrastive learning losses that are, respectively, based on offline and online association results. The offline association directly associates proxies according to the clustering and splitting results, while the online strategy dynamically associates proxies in terms of up-to-date features to reduce the noise caused by the delayed update of pseudo labels. The combination of two losses enables us to train a desirable Re-ID model. Extensive experiments on three person Re-ID datasets and one vehicle Re-ID dataset show that our proposed approach demonstrates competitive performance with state-of-the-art methods. Code will be available at: https://github.com/Terminator8758/O2CAP.
\end{abstract}


\section{Introduction}
Person re-identification (Re-ID) is the task of identifying the same person in non-overlapping cameras. Due to its significance in video surveillance and public security, this task has been extensively studied for decades. State-of-the-art performance is achieved mostly by supervised methods~\cite{chen2019abd,He2021vit}, requiring full labels that are expensive and time-consuming to annotate. Recently, semi-supervised~\cite{yangqi2019semi,Wang2021GCL} and unsupervised~\cite{lin2019aBottom,zhong2019invariance} Re-ID have been attracting more and more research interest, in a hope to reduce annotation cost and make the techniques more practical to real-world deployments. Although considerable progress has been achieved in these tasks, there is still a big gap in performance compared to the supervised counterparts. 

This work focuses on the purely unsupervised Re-ID task that requires no labels and thus fully releases the annotation burden. A promising research line for unsupervised Re-ID is clustering-based, which generates pseudo labels from the clustering results and uses the pseudo labels to learn a Re-ID model iteratively. With the popularity of contrastive learning~\cite{wu2018memory,Chen2020SimCLR,he2020momentum}, the integration of clustering and contrastive learning demonstrates great potential in recent unsupervised studies~\cite{ge2020self,Wang2021CAP,Chen2021ICE}. However, the pseudo labels generated by clustering is far from perfect, impeding Re-ID models from getting further improved. 

\begin{figure}[t]
\centering
\includegraphics[width=0.495\textwidth]{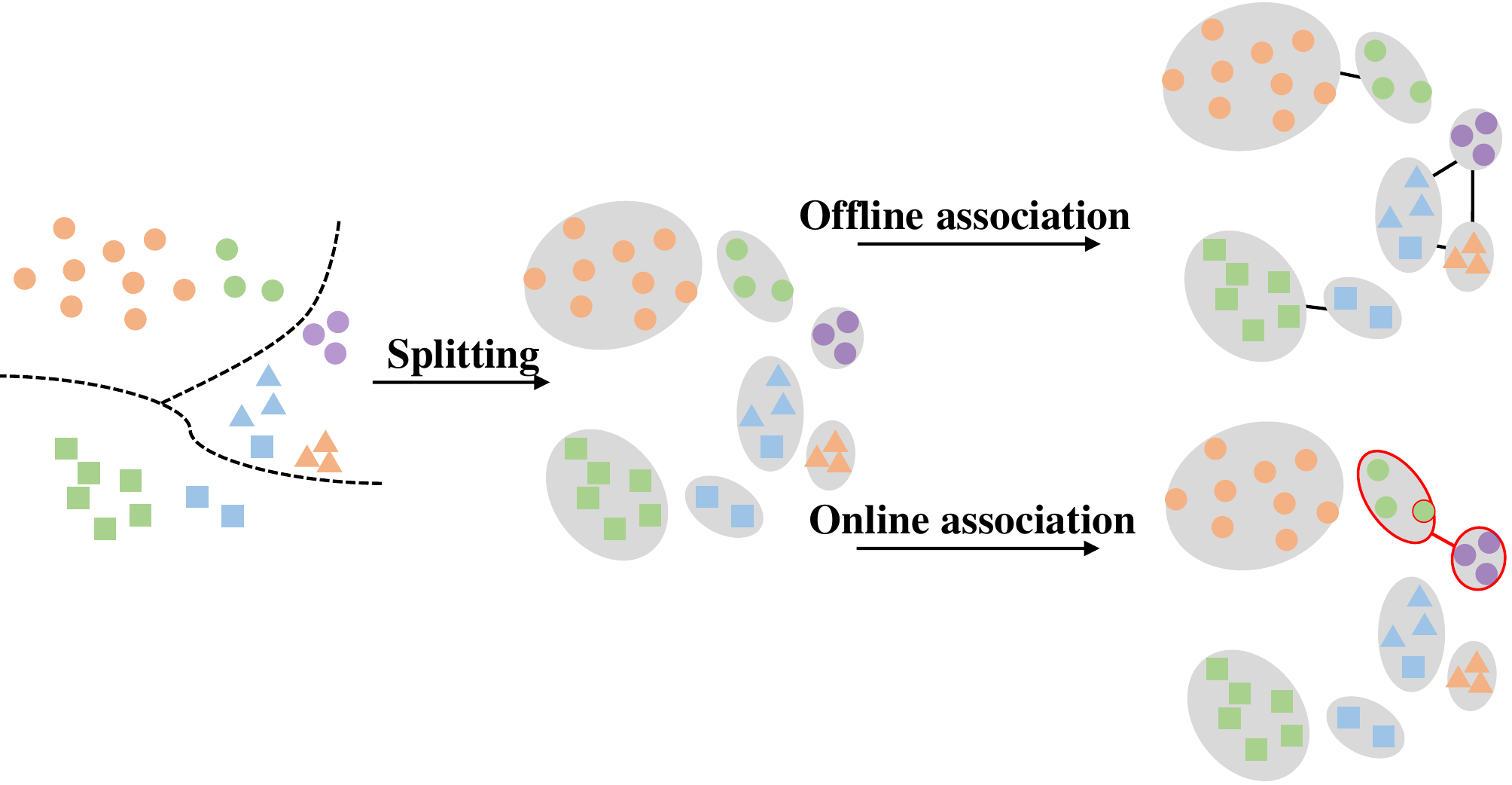} 
\caption{An illustration of the proposed camera-aware proxies, together with offline and online associations. Each cluster obtained by a camera-agnostic clustering step is split into multiple camera-aware proxies according to camera views. Proxies are further associated via an offline strategy and an online strategy to perform proxy-level contrastive learning. The offline strategy straightforwardly associates proxies split from the same cluster together. The online strategy dynamically associates positive proxies for an anchor instance (marked by red circles) in terms of up-to-date features. (Instances belonging to the same ID are represented by the same shape and those coming from the same camera are in the same color.)}
\label{fig_association}
\end{figure}

The noise of clustering-based pseudo labels mainly stems from two aspects. On one hand, pedestrian images are of high intra-ID variance and inter-ID similarity, which makes it extremely challenging for clustering algorithms such as DBSCAN~\cite{ester1996density} or K-Means to achieve accurate clusters. On the other hand, most methods perform clustering in an offline manner based on the features extracted at the beginning of each epoch, while instance features used for matching are extracted via a model updated on the fly. It implies that the generated pseudo labels might be out of date. To deal with the inevitable label noise, efforts have been made in label refinement~\cite{ge2020mutual,Wu2021MGH,zhang2021refine}, hybrid contrastive learning~\cite{ge2020self,Chen2021ICE,Sun2021} that combines cluster- and instance-level contrasts together, online pseudo label generation~\cite{Zheng2021}, and other techniques~\cite{Feng2021Pseudo,zeng2020hierarchical}. 

In this work, we propose camera-aware proxies to better deal with the intra-ID variance and inter-ID similarity. Our approach is inspired by the following observation. Since severe intra-ID variance is mainly caused by the change of camera views, an ID's intra-camera images tend to gather more tightly in a feature space than its inter-camera images. As a result, the clusters obtained by unsupervised clustering often present multiple sub-clusters, roughly corresponding to different camera views. Therefore, as shown in Figure~\ref{fig_association}, we propose to split each cluster, which is obtained by a camera-agnostic clustering method, into multiple camera-aware proxies according to camera views. Based on the pseudo labels that are generated from these camera-aware proxies, we design a proxy-level contrastive learning (CL) method that is more effective than cluster-level CL and requires less memory footprints than instance-level CL methods~\cite{ge2020self,Chen2021ICE,Sun2021}. 

To deal with the noise arisen from the offline clustering criterion as well as the delayed update of pseudo labels, we further propose a strategy that employs offline and online associations to complimentarily mine positive and hard negative proxies. Offline association is based on the offline clustering results. It straightforwardly associates the proxies split from a positive cluster as the positive ones. Online association aims to utilize up-to-date features of instances and proxies to associate positive proxies. To this end, we design an instance-proxy balanced similarity and a camera-aware nearest neighbor criterion to produce reliable associations, which are vital to make online association effective. Finally, two proxy-level contrastive learning losses are defined, respectively, according to the results of offline association and online association.

The main contributions are summarized as follows:
\begin{itemize}
	\item Instead of using camera-agnostic clusters, we propose camera-aware proxies that explicitly capture local structures within clusters. They enable us to perform proxy-level contrastive learning, which can tackle the intra-ID variance and inter-ID similarity better.
	\item With the assistance of the camera-aware proxies, we design a strategy that combines offline and online association based proxy-level contrastive learning together. This strategy can, to some extent, conquer the noise caused by the offline clustering criterion and the delayed update of pseudo labels. 
	\item Extensive experiments on three person Re-ID datasets and one vehicle Re-ID dataset show that the proposed method is competitive to state-of-the-art purely unsupervised and UDA-based Re-ID methods.
\end{itemize}

Note that this work is an extended version of our preliminary work CAP~\cite{Wang2021CAP}. In contrast to CAP, we make the following improvements:
\begin{itemize}
	\item This work proposes an online association scheme to utilize up-to-date features for proxy association, based on which an additional proxy-level contrastive learning loss is designed. 
	\item We discard the intra-camera contrastive learning loss that is used in CAP and analyze why this loss is not helpful to the new model in this work.
	\item More experiments on ablation studies and sensitivity analysis, as well as on a vehicle Re-ID dataset, are conducted to make a thorough validation. 
	\item Experiments show that this work improves CAP by a considerable margin on all datasets. Especially, on the most challenging dataset MSMT17, $4.6\%$ Rank-1 and $5.5\%$ mAP improvements are gained. 
\end{itemize}

%---------------------------------------------------------
\section{Related Work}
\subsection{Unsupervised Person Re-ID}
According to whether external labeled datasets are used or not, previous unsupervised methods can be classified into purely unsupervised and UDA-based groups.

\textit{Purely unsupervised person Re-ID} requires no annotations and therefore is more attractive. Existing methods commonly resort to pseudo labels for learning. Clustering~\cite{lin2019aBottom,zeng2020hierarchical}, k-NN~\cite{li2018unsupervised,chen2018deepa}, graph~\cite{ye2017dynamic,wu2019graph}, or hypergraph~\cite{Wu2021MGH} based techniques have been developed to generate pseudo labels. Clustering-based methods such as BUC~\cite{lin2019aBottom} and HCT~\cite{zeng2020hierarchical} conduct learning in a camera-agnostic way, which can capture the similarity within IDs but neglect the intra-ID variance mostly caused by the change of camera views. Alternatively, TAUDL~\cite{li2018unsupervised}, DAL~\cite{chen2018deepa}, UGA~\cite{wu2019graph}, and IICS~\cite{Xuan2021cvpr} divide the Re-ID task into intra- and inter-camera learning stages, by which the discriminative ability learned within cameras can further facilitate ID association across cameras. Our preliminary CAP~\cite{Wang2021CAP} proposes camera-aware proxies to deal with the intra-ID variance and conducts the unsupervised learning also from both intra- and inter-camera perspectives. But, in this extension work we find out that the intra-camera learning is not necessary, or even harmful, to perform effective learning when intra-camera pseudo labels are noisy. We therefore merely focus on the inter-camera learning while propose a combination of offline and online association to boost performance.

\textit{Unsupervised domain adaptation (UDA) based person Re-ID} demands a source dataset that is fully labeled, but leaves the target dataset unlabeled. To address this task, existing methods either transfer image styles~\cite{Wei2018PTGAN,Deng2018SPGAN,Liu2019ATNet,9018132} or reduce distribution discrepancy\cite{qi2019DA,Wu2019CA,Isobe2021} between different domains. These methods pay much attention to transfer knowledge from source domain to target domain. In addition, to sufficiently exploit unlabeled data in the target domain, clustering~\cite{unsup_clustering, zhai2020ad, ge2020self,Isobe2021,Wu2022} or k-NN~\cite{zhong2019invariance,9018132} based methods have also been adopted, analogous to those introduced in the purely unsupervised task. Differently, these methods either take both original and transferred data~\cite{unsup_clustering,zhong2019invariance,ge2020self,9018132} into account, or integrate a clustering procedure together with an adversarial learning step~\cite{zhai2020ad}. Although external labeled datasets are used, UDA-based methods do not gain noticeable advantage against recent purely unsupervised counterparts.

\textit{In both purely unsupervised and UDA-based person Re-ID}, early methods~\cite{li2018unsupervised,unsup_clustering,Wei2018PTGAN,Deng2018SPGAN} often use ID classification loss or triplet loss~\cite{hermans2017defense} for learning. Recently, contrastive learning (CL)~\cite{wu2018memory,Chen2020SimCLR,he2020momentum} has attracted a surge of research interest. Various CL-based methods~\cite{Wang2021CAP,Chen2021ICE,Liu2021,zhong2019invariance,ge2020self,Zheng2021,Isobe2021,9660560,Wu2022} have been developed for unsupervised Re-ID. Moreover, the issue of pseudo label noise occurred in unsupervised Re-ID has also been noticed recently and various methods~\cite{ge2020mutual,Zheng2021,zhang2021refine,Feng2021Pseudo,zeng2020hierarchical} have been proposed to address it. We will introduce these related work and state the difference of our work in Section~\ref{sec_labels} and~\ref{sec:cl}.

\subsection{Utilization of Camera Information}
As one type of valuable meta information, camera index has been extensively utilized in previous unsupervised Re-ID methods. For instance, CAMEL~\cite{yu2017cross} and UCDA-CCE~\cite{qi2019DA} aim to align distributions under different cameras by learning camera-specific projections or performing camera-aware domain adaptation. TAUDL~\cite{li2018unsupervised}, UTAL~\cite{li2019unsupervised}, UGA~\cite{wu2019graph}, IICS~\cite{Xuan2021cvpr}, and MGH~\cite{Wu2021MGH} combine intra-camera and inter-camera learning together to boost unsupervised Re-ID performance. Besides, we notice that camera information is also implicitly used in SpCL~\cite{ge2020self}. Although SpCL~\cite{ge2020self} does not explicitly use camera information in its model, it designs a MultiGallerySampler to ensure in-batch images of each ID to be sampled from diverse cameras, which enhances its performance further. These methods leverage camera information in various ways, but they do not pay attention to the intra-cluster variance caused by camera discrepancy. In this work, we propose camera-aware proxies to capture the camera-specific variance within each cluster, based on which offline and online association strategies are designed.

\subsection{Pseudo Label Refinement}
\label{sec_labels}
Clustering-based methods take a dominant role in recent unsupervised Re-ID. However, the pseudo labels generated by an offline clustering technique such as DBSCAN~\cite{ester1996density} inevitably contain noise. In order to reduce the label noise, various methods~\cite{ge2020mutual,Zheng2021,zhang2021refine,Feng2021Pseudo,zeng2020hierarchical} have been developed recently. For instance, RLCC~\cite{zhang2021refine} refines pseudo labels via clustering consensus over consecutive training generations. MGH~\cite{Wu2021MGH} adopts a hypergraph to propagate and refine pseudo labels generated by global DBSCAN, intra-camera DBSCAN, and global KNN. HCT~\cite{zeng2020hierarchical} utilizes hierarchical clustering and Zheng et al.~\cite{Zheng2021} propose an online clustering strategy to generate high quality pseudo labels. MMT~\cite{ge2020mutual} refines pseudo labels by the combined use of offline hard pseudo labels and online soft pseudo labels. In contrast to these methods, we propose camera-aware proxies (CAP) to conquer the clustering noise resulted from high complex data structures, and further design offline and online association strategies specific to CAP to deal with the noise stemmed from the clustering criterion as well as the delayed update of pseudo labels.

\subsection{Metric Learning with Proxies}
Metric learning plays an important role in person Re-ID and many other vision tasks. A loss extensively utilized in metric learning is the triplet loss~\cite{hermans2017defense}. It measures the distances of an anchor to a positive instance and a negative instance that are usually sampled within a batch. Along with the increment of the batch size, the number of triplets increases dramatically, resulting in a slow convergence and an inferior performance. To overcome this issue, Proxy-NCA~\cite{Attias2017proxy} and Center loss~\cite{CenterLoss16} propose to use proxies or centers, which represent sets of data instances, for the measurement of similarity and dissimilarity. The use of proxies instead of instances is able to capture more contextual information and greatly reduces the total number of triplets, enabling metric learning to achieve better performance. Further, with the awareness of intra-class variances, Magnet~\cite{Rippel2016multi-center}, MaPML~\cite{Qian2018proxy}, SoftTriple~\cite{qian2019softtriple}, GEORGE~\cite{Sohoni2020}, and ProxyGML~\cite{Zhu2020ProxyGML} adopt multiple proxies to represent a single cluster, by which local structures can be represented better. Our work is inspired by these studies. However, in contrast to set a fixed number of proxies for each class or design a complex strategy to find an appropriate proxy number, we split a cluster into a variant number of proxies simply according to the involved camera views, making our proxies more suitable for the Re-ID task.

\subsection{Contrastive Learning}
\label{sec:cl}
Contrastive learning (CL) performs learning also via comparing similarities of samples and therefore belongs to a special metric learning technique. In recent years, CL has attracted great interest due to its success in unsupervised representation learning tasks~\cite{wu2018memory,Chen2020SimCLR,he2020momentum}. 
Some losses such as Instance loss~\cite{InstanceLoss20} adopt a parametric form to conduct instance-level discrimination, while most typical contrastive losses such as InfoNCE~\cite{InfoNCE10,wu2018memory,he2020momentum,Chen2020SimCLR} perform in a non-parametric way, aiming to pull positive samples together while push negative samples apart. The losses are primitively focused on the contrast of instances and later extended to prototypes or proxies~\cite{Caron2020,Li2021pcl} as well. A crucial problem in CL is how to select positive and negative instances/proxies for effective comparison. Nowadays, this problem still remains open although various hard negative mining strategies~\cite{Robinson2021,Kalantidis2020} for contrastive learning have been proposed.

Recently, contrastive learning has also been widely applied to UDA-based~\cite{ge2020self,Zheng2021,Isobe2021,Wu2022} and purely unsupervised~\cite{Wang2021CAP,Chen2021ICE,Liu2021,9660560,groupsample21} person Re-ID tasks. For instance, SpCL~\cite{ge2020self} constructs a hybrid memory and executes contrastive learning at source-domain class-level, target-domain cluster-level, and un-clustered instance-level. Our CAP~\cite{Wang2021CAP} constructs a proxy-level memory bank to perform intra- and inter-camera contrastive learning at proxy-level. Later on, ICE~\cite{Chen2021ICE} and Liu et al.~\cite{Liu2021} boost SpCL and CAP via augmenting the models with instance-level contrastive learning, while MGH~\cite{Wu2021MGH} and Isobe \textit{et al.}~\cite{Isobe2021} integrate contrastive learning with hypergraph and Fourier augmentation. In contrast to them~\cite{Chen2021ICE,Liu2021,Sun2021,Wu2021MGH,Isobe2021,9660560}, this work extends CAP via sticking on the proxy-level contrastive learning alone to keep the approach simple yet effective.

\begin{figure*}[t]
\centering
\includegraphics[width=0.8\textwidth]{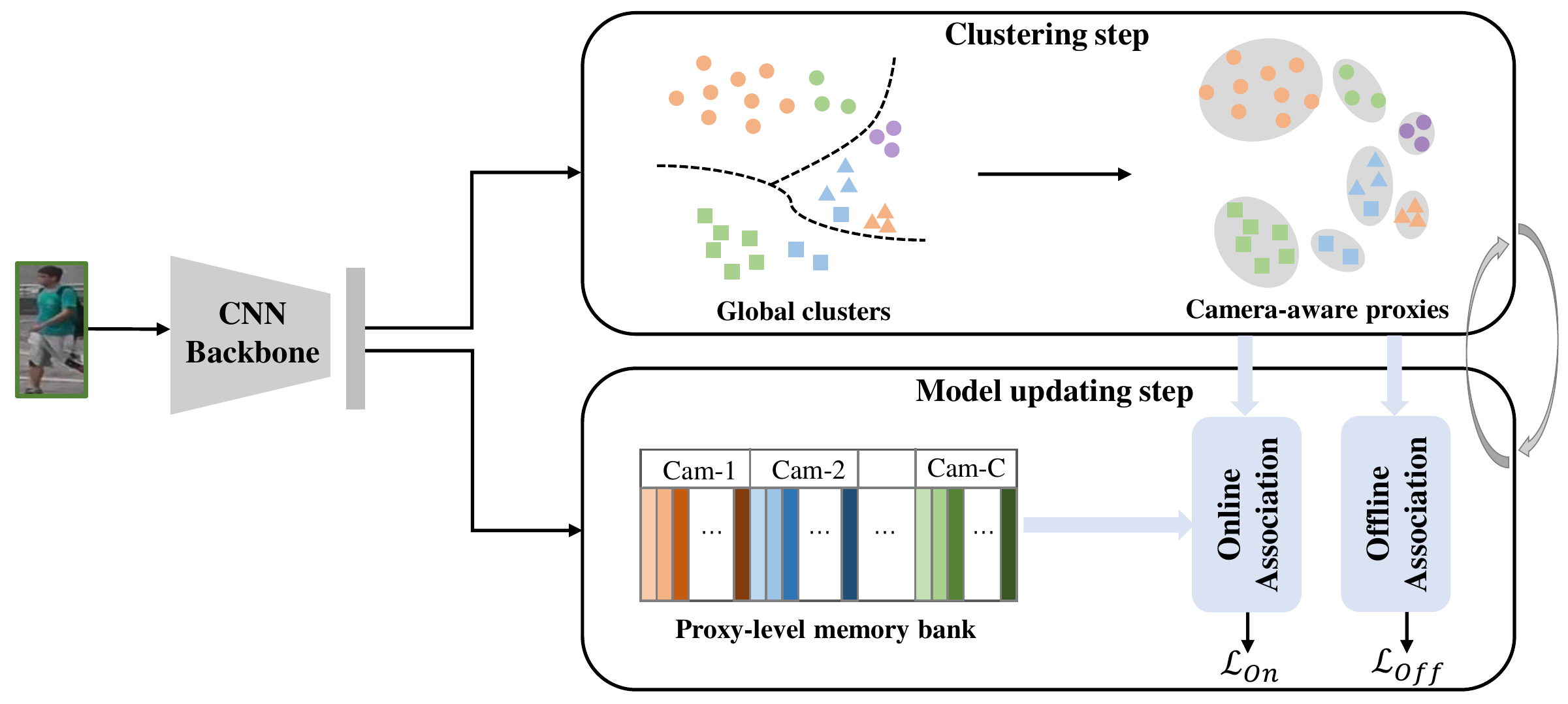} 
\caption{An overview of our proposed method. It iteratively alternates between a clustering step and a model updating step. In the clustering step, a global clustering is performed and then each cluster is split into multiple camera-aware proxies. In the model updating step, assisted with an external memory bank, two proxy-level contrastive learning losses $\mathcal{L}_{Off}$ and $\mathcal{L}_{On}$, which are respectively based on the results of offline and online associations, are optimized.
}
\label{fig_framework}
\end{figure*}

\section{A Clustering-based Re-ID Baseline}
\label{sec:baseline}
We first set up a baseline model for the purely unsupervised Re-ID task. As the common practice in clustering-based methods~\cite{unsup_clustering,lin2019aBottom,zeng2020hierarchical}, the baseline learns a Re-ID model iteratively and, at each epoch, it alternates between a clustering step and a model updating step. In contrast to previous methods that utilize an ID classification loss~\cite{unsup_clustering} or a triplet loss~\cite{zeng2020hierarchical}, we adopt a non-parametric Softmax loss~\cite{wu2018memory} for the model updating. This non-parametric loss, also termed as an InfoNCE loss~\cite{Oord2019CPC,he2020momentum}, plays an important role in recent contrastive learning techniques that have been successfully applied to various unsupervised learning tasks~\cite{wu2018memory,Chen2020SimCLR,he2020momentum,lin2019aBottom,zhong2019invariance,ge2020self}. It also makes our baseline model effective and extensible. In the followings, we briefly introduce the details. 

Given an unlabeled dataset $\mathcal{D} = \{x_i\}_{i=1}^{N}$, where $x_i$ is the $i$-th image and $N$ is the total number of images. We build the baseline model on a convolutional neural network (CNN) $f_\theta$ that is parameterized by $\theta$. The parameters are initialized from an ImageNet-pretrained~\cite{krizhevsky2012imagenet} network. When image $x$ is input, the network extracts a $d$-dimensional feature $f_\theta(x)$. Then, at each epoch, we adopt DBSCAN~\cite{ester1996density} to cluster the features of all images, and further select reliable clusters by simply discarding isolated outliers. All images within each cluster are assigned with a same pseudo ID label. By this means, we get a labeled dataset $\mathcal{D}' = \{(x_i, \tilde{y}_i)\}_{i=1}^{N'}$, in which $\tilde{y}_i \in \{1, \cdots, Y\}$ is a generated pseudo label. $N'$ is the number of images remained in the selected clusters and $Y$ is the number of clusters.

Once pseudo labels are generated, we adopt the contrastive learning technique for model updating. It is implemented via an external memory bank and an InfoNCE loss. Specifically, we construct a cluster-level memory bank $\mathcal{K}\in R^{d \times Y}$. During back-propagation, when image $x_i$ is input, we update the memory entry of its target ID class $\tilde{y}_i$ via a moving average scheme. That is,
\begin{equation}
	\mathcal{K}[\tilde{y}_i] \leftarrow \mu \mathcal{K}[\tilde{y}_i] + (1 - \mu) f_\theta(x_i),
	\label{eq:mu}
\end{equation}
where $\mathcal{K}[\tilde{y}_i]$ is the $\tilde{y}_i$-th entry of the memory bank, storing the updated feature centroid of class $\tilde{y}_i$, and $\mu \in [0,1]$ is an updating rate.

Then, the loss of the baseline model is defined by 
\begin{equation}
\mathcal{L}_{Base} = - \sum_{i=1}^{B} \log \frac{exp(\mathcal{K}[\tilde{y}_i]^T f_\theta(x_i)/\tau)}{\sum_{j=1}^{Y} exp(\mathcal{K}[j]^T f_\theta(x_i)/\tau)},
\label{eq_base}
\end{equation}
where $\tau$ is a temperature factor and $B$ is the batch size. This loss is a cluster-level contrastive learning loss, which aims to pull an instance close to the centroid of its class while push it away from the centroids of all other classes.

Considering that hard negative mining strategies may boost the performance of contrastive learning~\cite{Robinson2021,Kalantidis2020}, we additionally define an alternative loss of the baseline model as follows:
\begin{equation}
\mathcal{L}_{Base2} = - \sum_{i=1}^{B} \log \frac{exp(\mathcal{K}[\tilde{y}_i]^T f_\theta(x_i)/\tau)}{\sum_{j\in \{\tilde{y}_i\}\cup Q_0} exp(\mathcal{K}[j]^T f_\theta(x_i)/\tau)},
\label{eq_base2}
\end{equation}
where $Q_0$ contains the memory indexes of the hard negatives that are sampled via selecting the $K_1$-nearest negative clusters.

\section{The Camera-aware Proxy Assisted Method}
Like previous clustering-based methods~\cite{unsup_clustering,lin2019aBottom,zeng2020hierarchical,zhai2020ad}, the above-mentioned baseline model conducts the clustering and model updating steps in a camera-agnostic way. This way is able to maintain the similarity within each cluster, but may neglect the intra-cluster variance. Considering that severe intra-cluster variance is mainly caused by the change of camera views, we split each single cluster into multiple camera-specific proxies. Each proxy represents the instances coming from the same camera. 
The obtained camera-aware proxies provide us with an explicit way to deal with the variance within clusters. Besides, the proxies enable the contrastive learning to pay more attention to the hardest negative instances, which helps to reduce the inter-ID similarity. Therefore, with the assistance of the proxies, we design two inter-camera contrastive learning losses, which respectively take advantage of offline and online associated proxies, for the model updating. The entire framework is illustrated in Figure~\ref{fig_framework}, in which the modified clustering step and the improved model updating step are alternatively iterated.

More specifically, at each epoch, after the camera-agnostic clustering we split the clusters into camera-aware proxies, and generate a new set of pseudo labels that are assigned in a per-camera manner. That is, the proxies within each camera view are independently labeled. It also means that two proxies split from the same cluster may be assigned with two different labels. We denote the newly labeled dataset of the $c$-th camera by $\mathcal{D}_c = \{(x_i, \tilde{y}_i, \tilde{z}_i, c_i)\}_{i=1}^{N_c}$. Here, image $x_i$, which previously is annotated with a global pseudo label $\tilde{y}_i$, is additionally annotated with an intra-camera pseudo label $\tilde{z}_i \in \{1, \cdots, Z_c\}$ and a camera label $c_i = c \in \{1, \cdots, C\}$. $N_c$ and $Z_c$ are, respectively, the number of images and proxies in camera $c$, and $C$ is the camera number. Then, the entire labeled dataset is $\mathcal{D}''=\bigcup_{c=1}^C \mathcal{D}_c$.

Consequently, we construct a proxy-level memory bank $\mathcal{K}'\in R^{d \times Z}$, where $Z=\sum_{c=1}^C Z_c$ is the total number of proxies in all cameras. Each entry of the memory stores a proxy, which is updated by the same scheme as introduced in Eq. (\ref{eq:mu}) but in proxy-wise. With the proxies stored in the memory bank, we design offline and online association strategies to match the per-camera labeled proxies over all cameras, based on which two contrastive learning losses are proposed.

\begin{figure*}[t]
\centering
\includegraphics[width=0.85\textwidth]{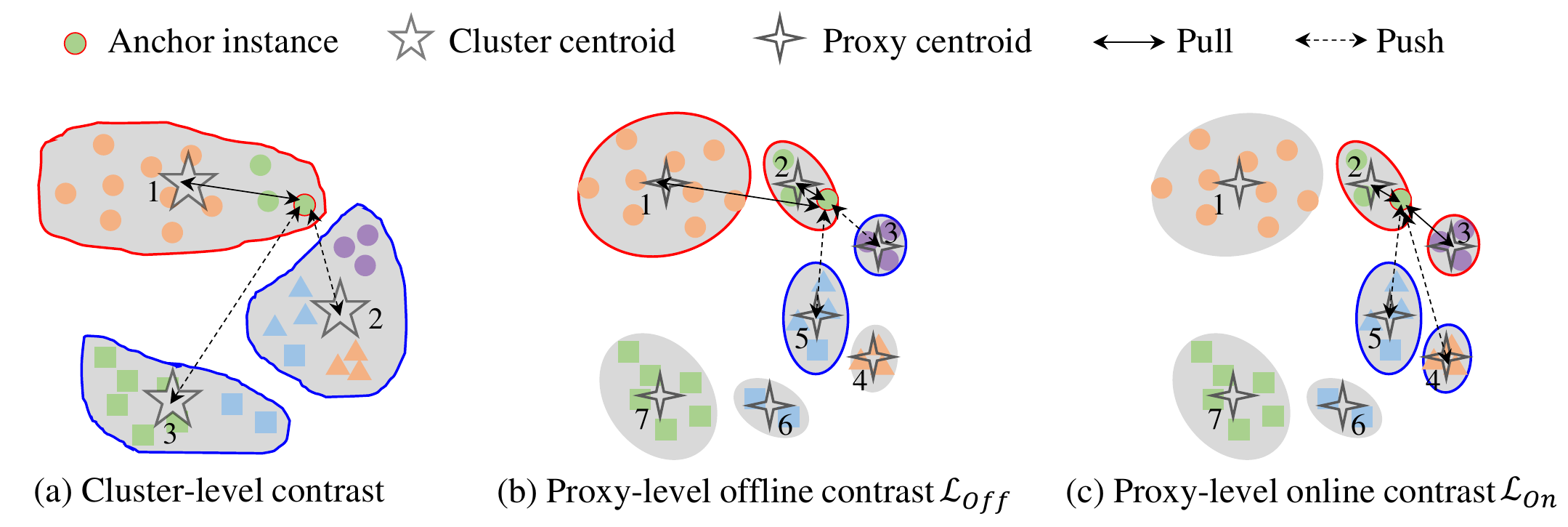} 
\caption{An illustration of different contrastive learning losses. An anchor instance and its positives are marked with red boundary, and its hard negatives are marked with blue boundaries. (a) The cluster-level contrast loss pulls the anchor instance close to the centroid of the cluster it belongs to, and pushes the anchor away from other cluster centroids. (b) The proxy-level offline contrast loss $\mathcal{L}_{off}$ pulls the anchor instance close to its positive proxies (e.g. Proxy 1 and 2) which are split from the positive cluster, and pushes the anchor away from the sampled hard negative proxies (e.g. Proxy 3 and 5). (c) The online contrast loss $\mathcal{L}_{on}$ pulls the anchor close to the positive proxies (e.g. Proxy 2 and 3) that are associated online, while repels it away from its hard negative proxies (e.g. Proxy 4 and 5). In this toy example, Proxy 3 is a false negative recalled by offline association but it is correctly associated as a positive proxy via online association.} 
\label{fig_contrast}
\end{figure*}

\subsection{The Contrastive Learning on Offline Associated Proxies}
We first design a contrastive learning loss according to the camera-agnostic clustering and camera-aware splitting that are conducted offline. As pointed out in~\cite{Robinson2021}, it is crucial to select appropriate positive and negative samples in order to perform effective contrastive learning. Fortunately, our camera-aware splitting strategy provides us with a straightforward way to find positive and negative proxies.

Specifically, given image $x_i$, we retrieve its positive proxies from all cameras, which share the same global pseudo label $\tilde{y}_i$. That is, all proxies split from the cluster $\tilde{y}_i$ are associated as positive ones. We refer to this association way as \textit{offline association} because the proxies are associated according to the results of offline clustering and splitting. The memory index set of these retrieved positive proxies is denoted by $\mathcal{P}_1$. Besides, we retrieve the $K_1$-nearest negative proxies from all remaining proxies as the hard negative ones, whose memory indexes are recorded by a set $\mathcal{Q}_1$. By this means, we define the first contrastive learning loss as follows.
\begin{equation}
\small
\mathcal{L}_{Off}=-\sum_{i=1}^{B}\left(\frac{1}{|\mathcal{P}_1|}\sum_{u \in \mathcal{P}_1} \log \frac{S(u, x_i)}{\sum\limits_{p \in \mathcal{P}_1} S(p, x_i) + \sum\limits_{q \in \mathcal{Q}_1} S(q, x_i)}\right),
\label{eq_inter1}
\end{equation}
in which $S(u, x_i) = exp (\mathcal{K}'[u]^T f_\theta(x_i) / \tau)$, and $|\mathcal{P}_1|$ is the cardinality of $\mathcal{P}_1$.

Note that this loss is an inter-camera contrastive learning term as both positive and negative proxies are retrieved across cameras. It in essence maximizes the multiplication of probabilities of $x_i$ being recognized as each positive proxy class. Thus, this loss pulls an instance close to all positive proxy centroids, which encourages a balanced learning for instance-rich and instance-deficient proxies within each cluster, leading to a high intra-ID compactness. Meanwhile, this loss also pushes the instance away from its hard negative proxies. In contrast to the cluster-level loss defined in Eq. (\ref{eq_base2}), the proxy-level hard negatives can capture local structures at a finer granularity and pay more attention to those hardest negative instances, as shown in Figure~\ref{fig_contrast} (b). Therefore, the proxy-level learning can conquer the inter-ID similarity better.

\subsection{The Contrastive Learning on Online Associated Proxies}
The contrastive learning loss defined above still suffers from noise. As shown in Figure~\ref{fig_contrast} (b), this loss may push false negatives away or pull false positives together due to the inaccurate clustering results. Considering that the inaccuracy arises from the density based clustering criterion (DBSCAN) as well as the out-of-date pseudo labels generated by offline clustering the features extracted at the beginning of each epoch, we propose an \textit{online association} strategy that utilizes the nearest neighbor criterion and up-to-date information for rectification. More specifically, it takes advantage of instance features extracted via the up-to-date Re-ID model together with the updated proxy entries to dynamically associate positive proxies for each anchor instance. In order to retrieve positive proxies more accurately, we design an \textit{instance-proxy balanced similarity} and a \textit{camera-aware nearest neighbor} criterion for association. Further, we define a contrastive learning loss based on the online association results.

\begin{figure}[h]
\centering
\includegraphics[width=0.45\textwidth]{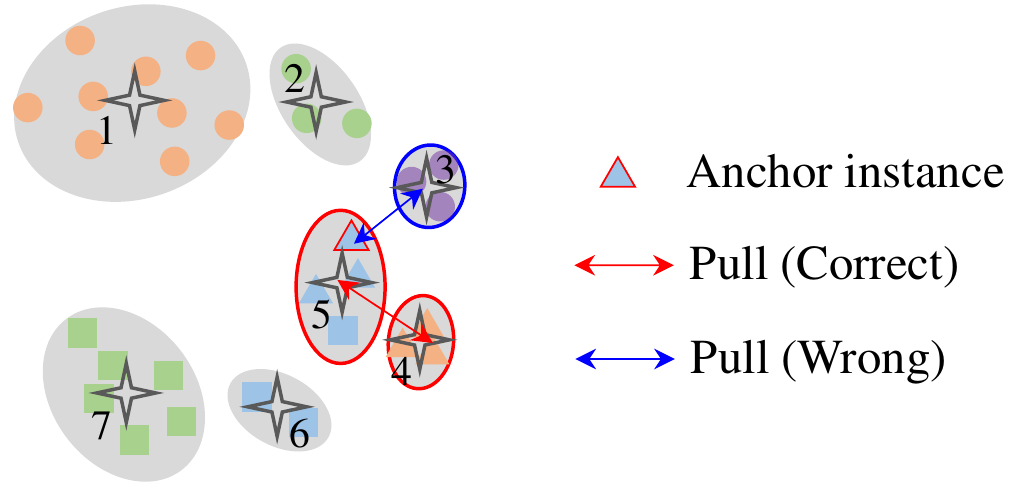}  
\caption{An example of the instance-proxy balanced similarity. For the anchor instance, the instance-to-proxy similarity tends to pull a false positive proxy (Proxy 3) close, while the proxy-to-proxy similarity pulls a true positive proxy (Proxy 4) together.}
\label{fig_sim}
\end{figure}

\textit{The instance-proxy balanced similarity} measures the similarity between an instance and a proxy based on the combination of an instance-to-proxy similarity and a proxy-to-proxy similarity. It is defined by
\begin{equation}
sim(f_\theta(x_i), \mathcal{K}'[j]) = w f^T_\theta(x_i)\mathcal{K}'[j]  + (1-w) \mathcal{K}'[\tilde{z}_i]^T \mathcal{K}'[j].
\label{eq_weighted_sim}
\end{equation}
Here, $\tilde{z}_i$ is the pseudo label of the proxy that $x_i$ belongs to, termed as the self-proxy, and $j$ is the index of any proxy. $w \in [0, 1]$ is a weight to balance the instance-to-proxy and proxy-to-proxy similarities. In contrast to a single instance-to-proxy similarity that is commonly used elsewhere, this balanced similarity is less sensitive to noise. A toy example is illustrated in Figure~\ref{fig_sim}. 
For the anchor instance marked with a red triangle, the instance-to-proxy similarity tends to associate a false proxy (Proxy 3) as a positive one, while the proxy-to-proxy similarity can associate a positive proxy (Proxy 4) correctly. The balanced similarity is helpful for such scenarios.

\textit{The camera-aware nearest neighbor criterion} is designed to associate at most one positive proxy within each camera. We propose this criterion based on the following observation: an ID's instances coming from the same camera are very likely to be grouped into the same cluster and thus into the same proxy. It implies that, for most anchor instances, there is at most one positive proxy existing in each camera. Therefore, directly using the global KNN, which tends to associate multiple proxies within one camera, may result in false positives. To reduce false positives, we propose the camera-aware nearest neighbor criterion. Specifically, given an instance image $x_i$, it first selects the 1-nearest neighbor proxy in each camera according to the instance-proxy balanced similarity, and then chooses Top-$K_2$ proxies from the selected ones as the positive proxies ($K_2$ is less than the number of cameras). The memory indexes of the associated positives are recorded in a set $\mathcal{P}_2$. In addition, we select $K_1$-nearest proxies from the remaining in terms of the instance-to-proxy similarity as the hard negatives and denote them by $\mathcal{Q}_2$.

Based on the online associated proxies, we define the second contrastive learning loss as 
\begin{equation}
\small
\mathcal{L}_{On}=-\sum_{i=1}^{B}\left(\frac{1}{|\mathcal{P}_2|}\sum_{u \in \mathcal{P}_2} \log \frac{S(u, x_i)}{\sum\limits_{p \in \mathcal{P}_2} S(p, x_i) + \sum\limits_{q \in \mathcal{Q}_2} S(q, x_i)}\right).
\label{eq_inter2}
\end{equation}

\subsection{A Summary of The Proposed Method}
The proposed method iteratively alternates between the camera-aware proxy clustering step and the contrastive learning based model updating step. The entire loss used for model updating is
\begin{equation}
\mathcal{L} = \mathcal{L}_{Off} + \mathcal{L}_{On},
\label{eq_overall_loss}
\end{equation}

To enable better understanding of our offline-online associated camera-aware proxies (O2CAP) based method, we summarize the overall procedure in Algorithm~\ref{our_algorithm}.

\begin{algorithm}[h]{
\caption{The O2CAP algorithm}
\label{our_algorithm}
\hspace*{0.02in} {\bf Input:}
Unlabeled training set $\mathcal{D}$, network $f_{\theta}$, number of training epochs \textit{maxEpoch}, number of batches \textit{numBatch} in one epoch, memory updating rate $\mu$, temperature $\tau$, and balancing weight $w$; \\
\hspace*{0.02in} {\bf Output:}
Trained network $f_{\theta}$;
\begin{algorithmic}[1]
\For{epoch = 1 to \textit{maxEpoch}}
    \State Perform a global clustering;
    \State Split clusters into camera-aware proxies;
    \State Construct proxy-level memory bank $\mathcal{K}'$;
    \State Generate the pseudo labeled dataset $\mathcal{D}''$;
    \For{b = 1 to \textit{numBatch}}
        \State Sample a mini-batch $\mathcal{B}' = \{(x_i, \tilde{y}_i, \tilde{z}_i, c_i)\}_{i=1}^{B}$;
        \State Obtain $\mathcal{P}_{1}$ and $\mathcal{Q}_{1}$ by offline association;
        \State Obtain $\mathcal{P}_{2}$ and $\mathcal{Q}_{2}$ by online association;
        \State Compute the loss $\mathcal{L}$ defined in Eq. (\ref{eq_overall_loss});
        \State Backward to update network parameter $\theta$;
        \State Update proxy entries in the memory $\mathcal{K}'$;
    \EndFor
\EndFor
\end{algorithmic}
}
\end{algorithm}

\textit{A proxy-balanced sampling strategy.} A mini-batch in Algorithm~\ref{our_algorithm} involves an update to the Re-ID model using a small set of samples. Apart from the loss optimization, the strategy of choosing appropriate samples in each batch is also important for model updating, especially when the data distribution is imbalanced~\cite{Mahajan2018,Zhang2021}. In this work, we propose a proxy-balanced sampling strategy that randomly chooses $P$ proxies and $K$ samples per proxy in each mini-batch. In contrast to the commonly used instance- or cluster-balanced sampling~\cite{hermans2017defense}, the proposed sampling strategy makes sure each proxy gets equal chance to be sampled. It thus facilitates the learning of image-deficient proxies, as well as the learning of the proxy-level contrastive losses.

\subsection{An Alternative of O2CAP} 
In the O2CAP model, we adopt two contrastive learning losses to, respectively, utilize the proxies obtained via offline and online association. Intuitively, there is another alternative way to utilize these offline and online associated proxies. That is, we get a positive proxy set via merging the offline and online associated positive proxies, i.e. $\mathcal{P}_3 = \mathcal{P}_1 \cup \mathcal{P}_2$. And then we select $K_1$-nearest proxies from the remaining as the hard negatives and denote them by $\mathcal{Q}_3$. By this means, instead of $\mathcal{L}$ defined in Eq. (\ref{eq_overall_loss}), we use the following alternative loss $\mathcal{L}_{Merge}$ for the model training, 
\begin{equation}
\small
\mathcal{L}_{Merge}=-\sum_{i=1}^{B}\left(\frac{1}{|\mathcal{P}_3|}\sum_{u \in \mathcal{P}_3} \log \frac{S(u, x_i)}{\sum\limits_{p \in \mathcal{P}_3} S(p, x_i) + \sum\limits_{q \in \mathcal{Q}_3} S(q, x_i)}\right).
\label{eq_alter}
\end{equation} 
However, in experiments we will show that $\mathcal{L}_{Merge}$ is inferior to the loss $\mathcal{L}$ that considers offline and online associated proxies separately.

\begin{table*}[!htp]
\caption{The statistics of each dataset. \texttt{\#}cameras, \texttt{\#}IDs, and \texttt{\#}images are the number of cameras, IDs, and images, respectively. CID is the averaged Camera-per-ID value and IID is the averaged Image-per-ID value.}
\centering
\scalebox{1}{
\begin{tabular}{c|ccccc|cc|cc}
\hline  %\toprule
\multirow{2}{*}{Dataset} & \multicolumn{5}{c|}{Training Set} &\multicolumn{2}{c|}{Gallery Set}&\multicolumn{2}{c}{Query Set}\\
\cline{2-10}  &  \texttt{\#}cameras  & \texttt{\#} IDs  &  \texttt{\#}images & CID & IID     & \texttt{\#}IDs  & \texttt{\#}images & \texttt{\#}IDs & \texttt{\#}images \\
\hline
Market-1501        & 6 & 751 & 12,936 & 4.34 & 17.23           & 751 & 15,913 & 750     & 3,368\\
DukeMTMC-reID      & 8 & 702 & 16,522  & 3.13 & 23.54         & 1,110  & 17,661 & 702 & 2,228  \\
MSMT17             & 15 & 1,041 & 32,621 & 4.63 & 31.34   	& 3,060 & 82,161  & 3,060  & 11,659  \\ \hline
VeRi-776           & 20 & 576 & 37,778 & 8.93  & 65.59							& 200 & 11,579 & 200 & 1,678 \\
\hline
\end{tabular}
}
\label{dataset_statistic_table}
\end{table*}

\subsection{Discussion on The Intra-camera Contrastive Learning}
All losses introduced above belong to inter-camera contrastive learning. In our preliminary CAP~\cite{Wang2021CAP}, an intra-camera contrastive learning loss is defined to learn the discriminative ability within cameras. We here make a brief introduction and discussion about it. Given image $x_i$, together with its per-camera pseudo label $\tilde{z}_i$ and camera label $c_i$, we set $A = \sum_{c=1}^{c_i-1} Z_c$ to be the total proxy number accumulated from the first to the $c_i-1$-th camera, and $j= A + \tilde{z}_i$ to be the index of the corresponding entry in the proxy-level memory. Then, the intra-camera contrastive learning loss is defined by 
\begin{equation}
\small
\mathcal{L}_{Intra} = -\sum_{c=1}^{C}\left(\frac{1}{N_c}\sum_{x_i\in \mathcal{D}_c} \log \frac{exp(\mathcal{K}'[j]^T f_\theta(x_i)/\tau)}{\sum_{k=A+1}^{A+N_{c_i}} exp(\mathcal{K}'[k]^T f_\theta(x_i)/\tau)} \right).
\label{eq_memory_prob}
\end{equation}
This loss performs contrastive learning within each camera. It pulls an instance close to the proxy to which it belongs, while pushes away from all other proxies in the same camera and ignores the proxies in other cameras.

The combination of this loss with $\mathcal{L}_{Off}$ (denoted as $\mathcal{L}_{Inter}$ in CAP~\cite{Wang2021CAP}), together with the proxy-balanced sampling strategy, enable CAP to gain the best performance among its model variants. However, during experiments we observe that adding this intra-camera loss to the O2CAP model is not necessary any more. Considering that the per-camera pseudo labels generated by camera-aware proxies inevitably contain noise, we conjecture that this loss is more effective when a Re-ID model is relatively weak. When O2CAP has already achieved great discriminative ability by the synergy of offline and online association, the intra-camera loss learned from noisy labels may bring confusion to the model and lead to a performance degeneration. Therefore, we discard this loss in our O2CAP model.

\section{Experiments}
\subsection{Datasets and Evaluation Metrics}
We evaluate the proposed method on three person Re-ID datasets: Market-1501~\cite{7410490}, DukeMTMC-reID~\cite{ristani2016performance,zheng2017unlabeled}, and MSMT17~\cite{Wei2018PTGAN}. The Market-1501 and DukeMTMC-reID datasets are collected on university campus, containing outdoor scenarios only. The MSMT17 dataset is a larger and more challenging dataset, which contains both indoor and outdoor scenarios under different weather conditions. In order to validate the generalization ability of our method, we additionally evaluate it on a vehicle Re-ID dataset, VeRi-776~\cite{liu2018veri}. The statistics of these datasets, including the number of cameras, IDs, and images contained in training, gallery, and query sets, are summarized in Table~\ref{dataset_statistic_table}. Meanwhile, the averaged Camera-per-ID (CID) value and the averaged Image-per-ID (IID) value are also provided for reference. 

For performance evaluation, we adopt the commonly used mean Average Precision (mAP) and Cumulative Matching Characteristic (CMC) as the metrics. The CMC metric is reported via Rank-1, Rank-5, and Rank-10. To make a fair comparison, we do not use any post-processing techniques (e.g. Re-ranking~\cite{Zhong2017reranking}) during evaluation.

\subsection{Implementation Details}
We adopt a slightly modified ResNet-50~\cite{he2016deep} as the network backbone. The modifications are as follows: we discard the classification layer in ResNet-50 but additionally add a Batch Normalization (BN) layer right after the Global Average Pooling (GAP) layer, following~\cite{hermans2017defense, 8237672, luo2019trick}. The scale and shift parameters of the additional BN layer are initialized as 1 and 0 respectively. The remaining parameters of the backbone are initialized from the original ResNet-50 that is trained on ImageNet. The BN layer outputs a 2048-dimensional feature for each image. Each feature is further normalized by $L_2$ norm and then used for the computation of the losses as well as the update of memory entries during training. The $L_2$ normalized features are also used for the distance computation during test time.

The hyper-parameters involved in our model are empirically set as follows. The memory updating rate $\mu$ is $0.2$ and the temperature factor $\tau$ in all contrastive losses is set to $0.07$. The number of hard negatives, i.e. $K_1$, is fixed to $50$. The balancing weight $w$ in Eq. (\ref{eq_weighted_sim}) is $0.15$. The number of positive proxies associated online, i.e. $K_2$, is set to 3 for both Market-1501 and MSMT17, 2 for DukeMTMC-reID, and 8 for VeRi-776, roughly close to but less than their CID values listed in Table~\ref{dataset_statistic_table}. At the beginning of each epoch, we compute Jaccard distance~\cite{Zhong2017reranking} of all features and then use DBSCAN~\cite{ester1996density} with a threshold of $0.5$ and neighborhood $eps=4$ to conduct the camera-agnostic clustering.

In addition, we adopt the commonly used random flipping, cropping, and erasing to augment data for training. The model is trained by ADAM~\cite{Kingma2014} optimizer with $\beta_1 = 0.9$, $\beta_2 = 0.999$, and weight decay of $0.0005$. The learning rate is initially set to 0.00035 with a warmup in the first 10 epochs, and is divided by 10 after each 20 epochs. The number of total epochs is 50. The number of iterations in each epoch is set to 400, as the common practice~\cite{ge2020mutual,ge2020self,Chen2021ICE}. The batch size is 32. Following the proposed proxy-balanced sampling strategy, we randomly sample 8 camera-aware proxies and 4 images per proxy within each batch. Our model is implemented with the Pytorch~\cite{PyTorch} framework. All experiments are run on a single GTX 1080Ti GPU. The training phase (50 epochs) takes about 2.5 hours for Market-1501 and DukeMTMC-reID, 3.5 hours for MSMT17, and 4 hours for VeRi-776 dataset.

\begin{table*}[t]
\centering
\caption{Comparison of the proposed method and its variants. \textit{PBsampling} is the proxy-balanced sampling strategy. When \textit{PBsampling} is not selected, the models use the conventional cluster-balanced sampling strategy. Note that $\mathcal{L}_{Off}$ is the same as $\mathcal{L}_{Inter}$ in CAP~\cite{Wang2021CAP}. And due to re-implementation with some hyper-parameters altered (e.g. iteration number per epoch), the performance values of CAP and its variants are not exactly the same with those reported in the preliminary work~\cite{Wang2021CAP}.}
\scalebox{1.05}{
\begin{tabular}{c|cccc|cc|cc|cc|cc}
\hline  %\toprule
\multirow{2}{*}{Models} & \multicolumn{4}{c|}{Components} &\multicolumn{2}{c| }{Market-1501} & \multicolumn{2}{c|}{DukeMTMC-ReID}  & \multicolumn{2}{c|}{MSMT17} & \multicolumn{2}{c}{VeRi-776}  \\
\cline{2-13}  &$\mathcal{L}_{Intra}$ & $\mathcal{L}_{Off}$ &$\mathcal{L}_{On}$ & \textit{PBsampling} & R1  & mAP   & R1  & mAP   & R1 & mAP & R1  & mAP \\ 
\hline
 Baseline1 & \multicolumn{4}{c|}{$\mathcal{L}_{Base}$} &   84.4 & 67.7       & 65.8  & 48.1            & 29.4  & 13.0    & 46.2 & 18.0 \\ 
 Baseline2 & \multicolumn{4}{c|}{$\mathcal{L}_{Base2}$}&   84.9 & 68.9    & 72.2  & 55.3            & 34.3  & 15.3    & 52.0 & 19.4  \\ \hline
CAP1 & \checkmark &  &  &   & 76.4 & 57.5          & 74.2  & 57.4            & 43.7 & 20.3                 & 76.4  & 35.1  \\
CAP2 & \checkmark &  &  &\checkmark    & 83.6 & 67.1      & 76.5  & 60.8           & 42.1 & 18.3                  & 79.1  & 36.5   \\
CAP3 &  & \checkmark &  & & 89.3  & 76.1        & 77.1  & 61.1         & 63.7  & 33.2             & 78.7 & 33.6\\
CAP4 &  & \checkmark & & \checkmark  & \textbf{92.5} & \textbf{82.8}           & 79.6  & 64.0          & 66.6  & 36.9   & 80.6  & 35.5\\
CAP5 & \checkmark &  \checkmark & &  & 90.1 & 78.7      & 77.3  & 61.2           & 61.6 & 32.2                  & 82.7  & 39.5   \\
\rowcolor{mygray}
\textbf{CAP} & \checkmark & \checkmark & & \checkmark     & 92.2  & 81.6        & 81.2  & 68.3          & 65.9  & 35.5       & 83.8  & 38.7 \\ 
\hline 
O2CAP1 & \checkmark & \checkmark &\checkmark & \checkmark & 92.0  & 82.0     & 82.0  & 69.0          & 71.3  & 41.7        & 87.1  & 41.3\\
\rowcolor{mygray}
\textbf{O2CAP} & & \checkmark &\checkmark & \checkmark &  \textbf{92.5} & 82.7       & \textbf{83.9}  & 71.2        & \textbf{72.0}  & \textbf{42.4}       & \textbf{87.5}  & \textbf{41.9}\\
O2CAP2 & & &\checkmark & \checkmark &  91.4 & 80.2       & 82.0  & 68.6          & 57.2  & 28.2       & 80.9  & 35.2   \\
O2CAP3 & & \multicolumn{2}{c}{$\mathcal{L}_{Merge}$}  & \checkmark & 92.1  & 81.8           & 83.7   & \textbf{71.3}            & 67.5 & 38.3        & 85.2 & 38.9\\   
\hline
\end{tabular}
}
\label{ablation_table}
\end{table*}

\subsection{Ablation Studies}
In this subsection, we conduct a series of experiments to validate the effectiveness of each proposed component. The performance of our full model and its variants are presented in Table~\ref{ablation_table}.

\subsubsection{Effectiveness of the proxy-level contrastive learning} Let us first compare CAP3 with Baseline2. The CAP3 model (using $\mathcal{L}_{Off}$ only) and the Baseline2 model (using $\mathcal{L}_{Base2}$) perform the contrastive learning, respectively, at the proxy-level and cluster-level while keep other settings the same. They pull an instance, respectively, close to its positive proxies or its positive cluster. Since offline association straightforwardly associates the proxies split from one positive cluster as the positive proxies, both models in essence encourage high compactness within the same clusters. However, as shown in Figure~\ref{fig_contrast}, the proxy-level learning achieves a more balanced learning of instance-rich and instance-deficient proxies, leading to a higher intra-cluster compactness. The other difference of these two models lies in that CAP3 pushes an instance away from its \textit{hard negative proxies} while Baseline2 pushes the instance away from its \textit{hard negative clusters}. Therefore, CAP3 pays more attention to the hardest negative instances so that the inter-ID similarity is better conquered. Last but not least, the pseudo labels generated from proxies are less noisy than those obtained from clusters, which also benefit the model learning. Due to these reasons, CAP3 outperforms Baseline2 by a great margin, especially on the complex datasets such as MSMT17 and VeRi-776, as shown in Table~\ref{ablation_table}.

\subsubsection{Effectiveness of the combination of offline and online association} 
From O2CAP1 vs. CAP and O2CAP vs. CAP4, we observe that the models additionally integrated with the online association based loss ($\mathcal{L}_{On}$) can boost the performance by a considerable margin on the datasets including DukeMTMC-ReID, MSMT17, and VeRi-776. Besides, the combination of offline and online association in O2CAP is implemented via the sum of two contrastive losses $\mathcal{L}_{Off}$ and $\mathcal{L}_{On}$. To validate the effectiveness of this combination way, we compare it with an alternative way that directly merges the positive proxy sets obtained offline and online into one set and adopts one loss $\mathcal{L}_{Merge}$ to perform contrastive learning. As shown in Table~\ref{ablation_table}, the O2CAP model (using $\mathcal{L}_{Off}$ and $\mathcal{L}_{On}$) performs comparable with the O2CAP3 model (using $\mathcal{L}_{Merge}$) on Market-1501 and DukeMTMC-ReID, but significantly outperforms O2CAP3 on MSMT17 and VeRi-776, demonstrating the superiority of the two-loss combination way in complex scenarios.

\begin{figure*}[t]
\centering
\subfloat[]{\includegraphics[width=0.31\textwidth]{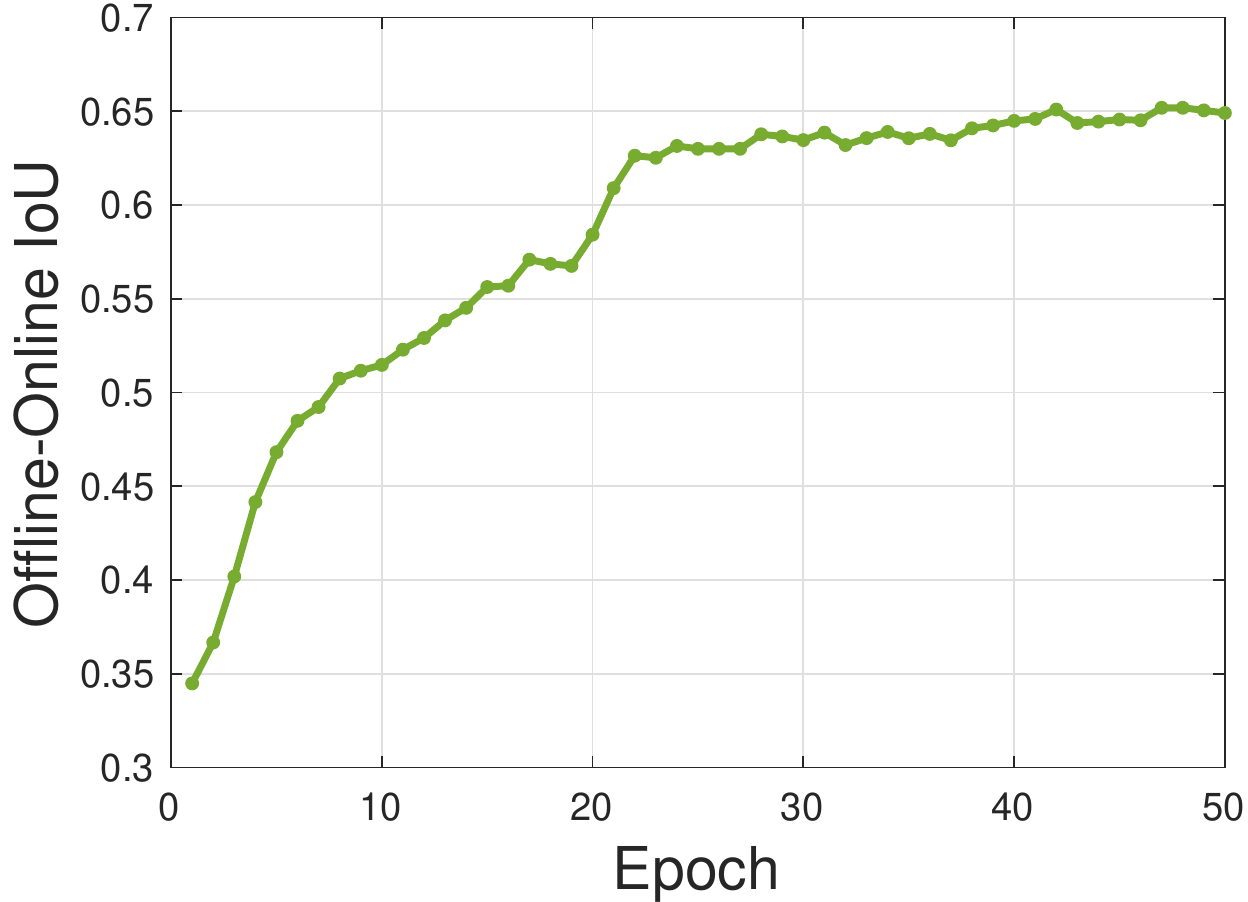}}
\quad
\subfloat[]{\includegraphics[width=0.315\textwidth]{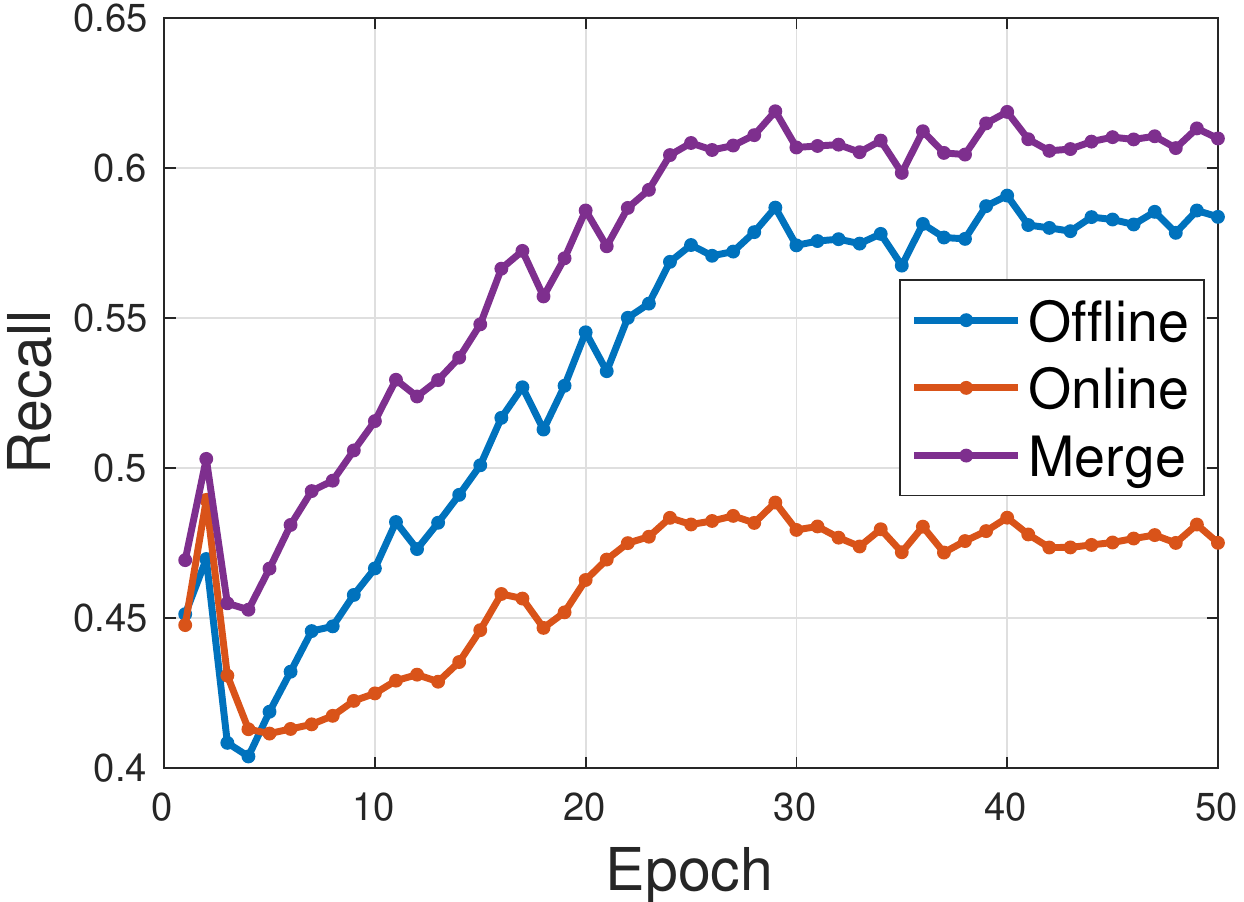}}
\quad
\subfloat[]{\includegraphics[width=0.305\textwidth]{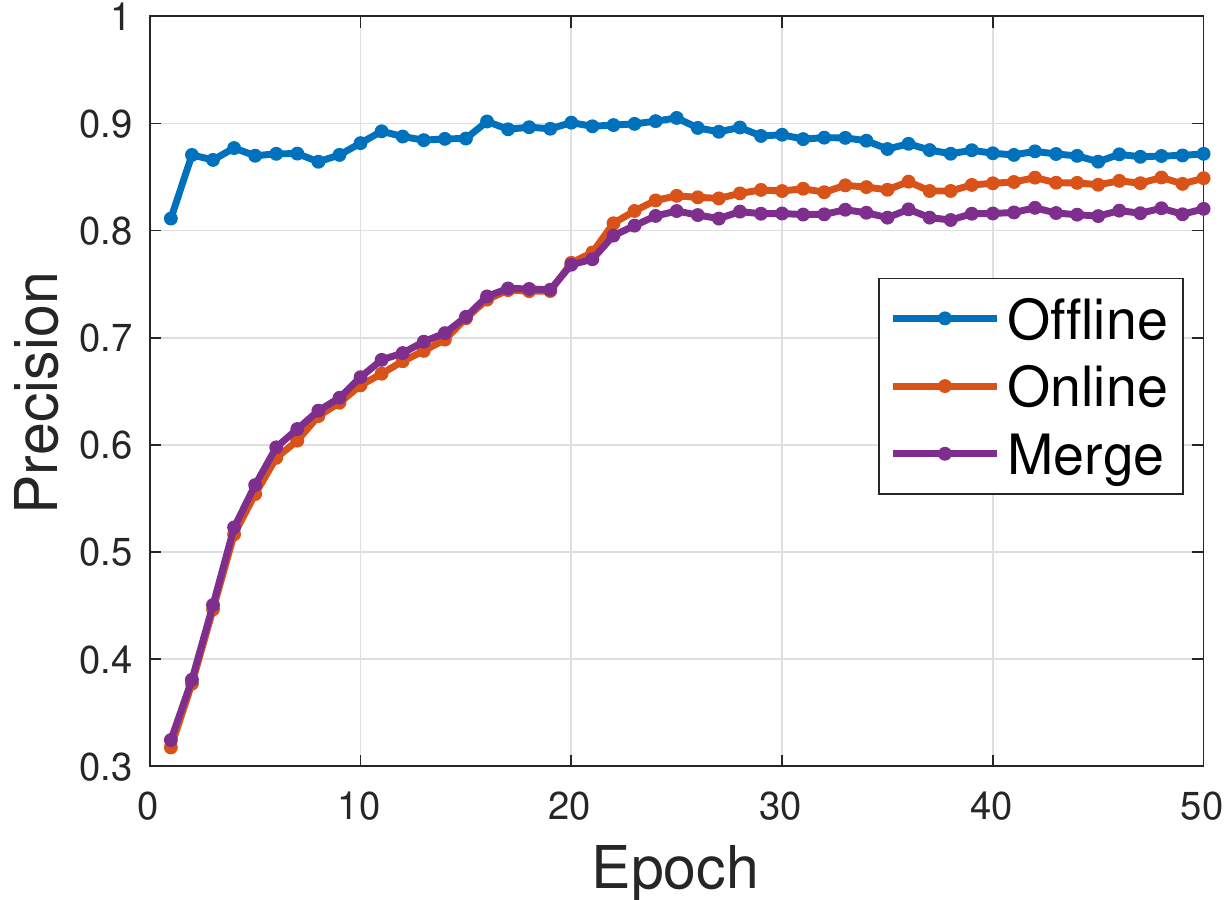}}
\caption{The statistics of associated positive proxies during training on MSMT17. (a) is the averaged IoU of offline and online associated proxies for each instance. (b) includes the averaged recall of offline, online associated positive proxies and their union. (c) includes the averaged precision of offline, online associated positive proxies and their union.}
\label{fig_analyze_association}
\end{figure*}

To further analyze the complementarity of offline and online association, we investigate some typical association statistics. Specifically, the offline and online proxy sets associated to each anchor instance $x_i$ are denoted as $\mathcal{P}_1^i$ and $\mathcal{P}_2^i$, respectively, and the union set is $\mathcal{P}_3^i = \mathcal{P}_1^i \cup \mathcal{P}_2^i$. We further assign each proxy with a ground-truth ID label via taking the ID of its majority instances, by which the ground-truth positive proxy set $\mathcal{P}_{gt}^i$ of $x_i$ can be roughly obtained. Then, the averaged Intersection over Union (IoU) of offline and online associated proxies, together with the averaged precision and recall of the proxies associated in different ways, are computed as followings:
\begin{eqnarray}
\small
IoU &=& \frac{1}{N'} \sum_{i=1}^{N'}\frac{|\mathcal{P}_1^i \cap \mathcal{P}_2^i|}{|\mathcal{P}_1^i \cup \mathcal{P}_2^i|}, \\
precision &=&  \frac{1}{N'} \sum_{i=1}^{N'} \frac{|\mathcal{P}_r^i \cap \mathcal{P}_{gt}^i|}{|\mathcal{P}_r^i|}, \\
recall &=& \frac{1}{N'} \sum_{i=1}^{N'} \frac{|\mathcal{P}_r^i \cap \mathcal{P}_{gt}^i|}{|\mathcal{P}_{gt}^i|}
\end{eqnarray}
in which $N'$ is the total training images and $r = 1, 2,$ or $3$ denotes the offline, online, or union association set.

Figure~\ref{fig_analyze_association} plots the detailed association statistics varying during training on MSMT17. As shown in Figure~\ref{fig_analyze_association} (a), the IoU between offline and online associated proxies increases from an initial value of $0.34$ to a converged value of $0.65$. It indicates that the two associations are getting more consistent as training goes on, but a part of them are still unique, leading to their complementarity and synergy. Figure~\ref{fig_analyze_association} (b) and (c) present the recall and precision of two association schemes. An intuitive observation is that the precision and recall of two associations are improving during training, and offline association achieves higher performance than online association. Nevertheless, the union of offline and online associations generates higher recall than either alone, showing that offline and online association indeed complement each other, and together they retrieve more positive proxies to benefit model learning. On the other hand, the precision of the union association is lower than offline or online association, which may explain the compromised performance of O2CAP3.

\subsubsection{Effectiveness of the strategies in online association}
In the design of our online association, we propose an instance-proxy balanced similarity and a camera-aware nearest neighbor criterion to select positive proxies. In order to validate their effectiveness, we compare them, respectively, with the original instance-to-proxy similarity and the global KNN that are ordinarily used~\cite{zhong2019invariance,Wu2021MGH}. Table~\ref{ablation_online_association2} presents the comparison results. Compared to the global KNN, the camera-aware criterion can avoid to associate multiple proxies within one camera and therefore reduce false associations. It plays a vital role to make online association effective. From Table~\ref{ablation_online_association2} and Table~\ref{ablation_table} we see that, when the global KNN is adopted, the online association is sensitive to the similarity measurement and is not always leading to performance enhancement. On the contrary, the models adopting the camera-aware criterion consistently improve performance no matter which similarity is used. When the camera-aware criterion is adopted, the balanced similarity can boost the performance further. The best performance is achieved when the balanced similarity is used for the association of positive proxies while the instance-to-proxy similarity is used for negative mining. 

\begin{table}[h]
\centering
\caption{Comparison of the online association strategies and their counterparts. Experiments are conducted on the O2CAP model. Global KNN chooses K-nearest neighbors globally and `CA' is our camera-aware criterion. For the similarity measurements, `Balanced' refers to the balanced similarity and `Original' is the single instance-to-proxy similarity.}
\scalebox{1}{
\begin{tabular}{ccc | cc | cc} 
\hline
\multicolumn{3}{c|}{Settings} &  \multicolumn{2}{c|}{DukeMTMC-reID}  & \multicolumn{2}{c}{MSMT17}\\
\hline
KNN & Positives & Negatives         & R1  & mAP            & R1  & mAP  \\ 
 \hline
Global            &  Original         & Original       & 80.7 & 67.2             & 69.1 & 39.7    \\
Global            &  Balanced & Balanced      &  81.3  & 66.9          & 60.0 & 32.2  \\
Global            &  Balanced & Original      &  80.6  & 66.3            & 50.0 & 24.2  \\
\hline
CA  &  Original         & Original      & 82.8 & 70.7              & 68.9 & 39.2    \\
CA  &  Balanced & Balanced     &  83.4  & 69.2          & 71.1 & 41.1  \\
\rowcolor{mygray}
CA  &  Balanced & Original    & \textbf{83.9}  & \textbf{71.2}        & \textbf{72.0}  & \textbf{42.4} \\
\hline
\end{tabular}
}
\label{ablation_online_association2}
\end{table}

\subsubsection{Effectiveness of the proxy-balanced sampling strategy}
In order to validate the effectiveness of the proposed proxy-balanced sampling strategy, we compare it to the commonly used cluster-balanced strategy that takes no consideration of intra-cluster distributions. From CAP2 vs. CAP1, CAP4 vs. CAP3, and CAP vs. CAP5, we observe that the models using the proxy-balanced sampling strategy consistently outperform the counterparts using the cluster-balanced strategy on almost all datasets. The results show that the proxy-balanced sampling strategy facilitates the learning of all camera-specific proxies and naturally fits to the proxy-level contrastive learning.

\subsubsection{Analysis on the intra-camera contrastive learning} The intra-camera contrastive learning loss ($\mathcal{L}_{Intra}$) was proposed in CAP~\cite{Wang2021CAP} to take advantage of the per-camera pseudo labels to learn discriminative ability within cameras and further boost the learning of global discrimination. As shown by CAP1 and CAP2, this loss enables the models to gain considerable discriminative ability. And from the full model CAP, we see that the integration of $\mathcal{L}_{Intra}$ and $\mathcal{L}_{Off}$ does improve the performance over the model CAP4 that utilizes $\mathcal{L}_{Off}$ only. However, when comparing O2CAP with O2CAP1, we find out that the integration of $\mathcal{L}_{Intra}$, on the contrary, degrades the performance on all datasets. Our conjecture is as follows. The per-camera pseudo labels generated from camera-aware proxies are, although more reliable than the cluster-level pseudo labels, still noisy. Thus, the noisy label based intra-camera learning may bring confusion to the O2CAP model that has already achieved great discriminative ability.

\subsection{Parameter Analysis}
In this subsection, we analyze the sensitivity of the hyper-parameters involved in O2CAP. Among all hyper-parameters, the memory updating rate $\mu$ and the temperature factor $\tau$ have been investigated in many other works so that we simply follow~\cite{Wang2021GCL,Wang2021CAP} to set them. Here, we conduct experiments on DukeMTMC-reID and MSMT17 to investigate the sensitivity of the remaining hyper-parameters, which include the number of hard negative proxies ($K_1$), the number of positive proxies associated online ($K_2$), and the weight $w$ in the instance-proxy balanced similarity. In addition, the sensitivity of the training epochs is also investigated as a reference.

\subsubsection{The sensitivity of $K_1$} 
\begin{figure}[h]
\centering
\subfloat[\scriptsize{DukeMTMC-reID}]{\includegraphics[width=0.225\textwidth]{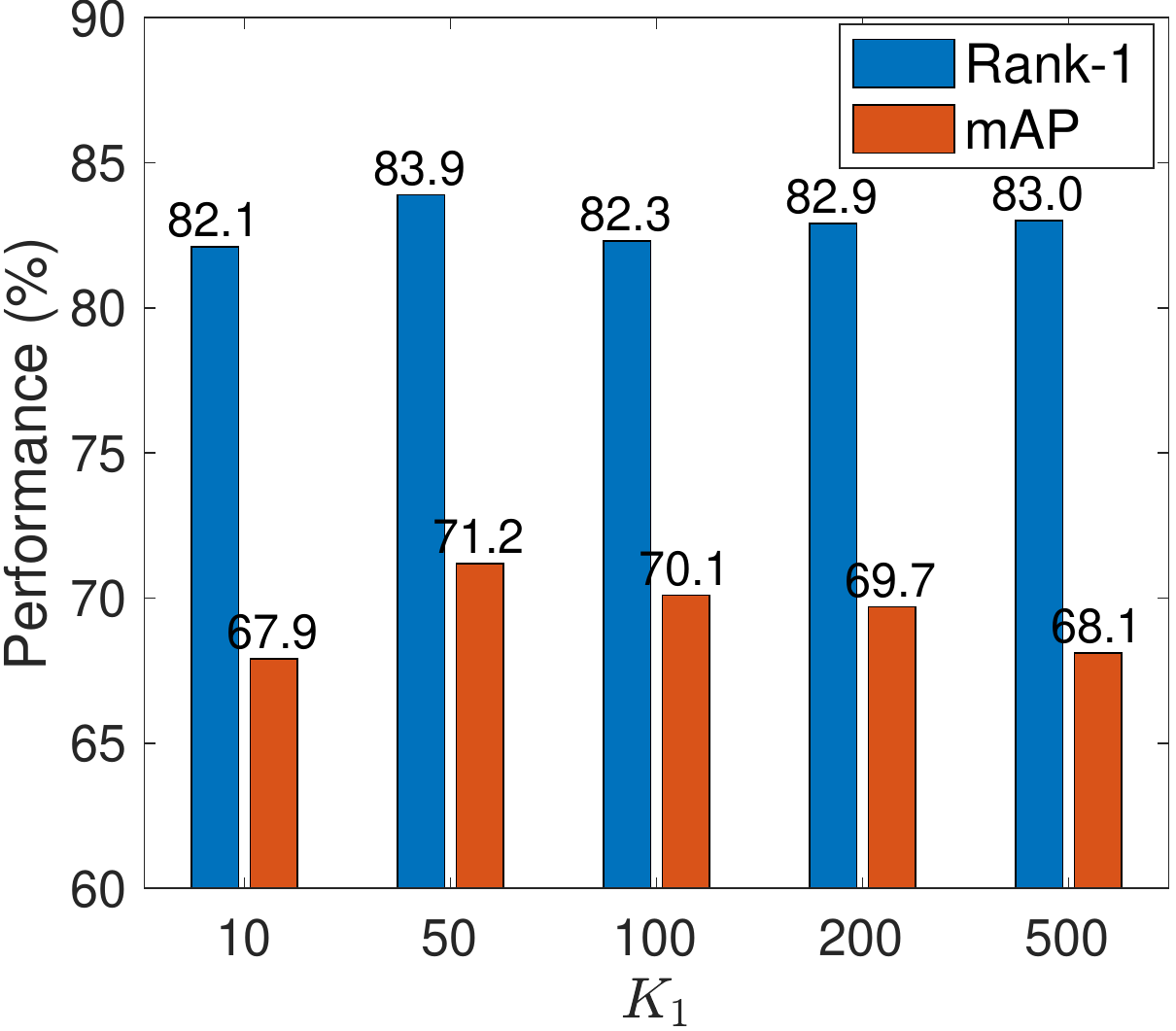}}
\quad
\subfloat[\scriptsize{MSMT17}]{\includegraphics[width=0.225\textwidth]{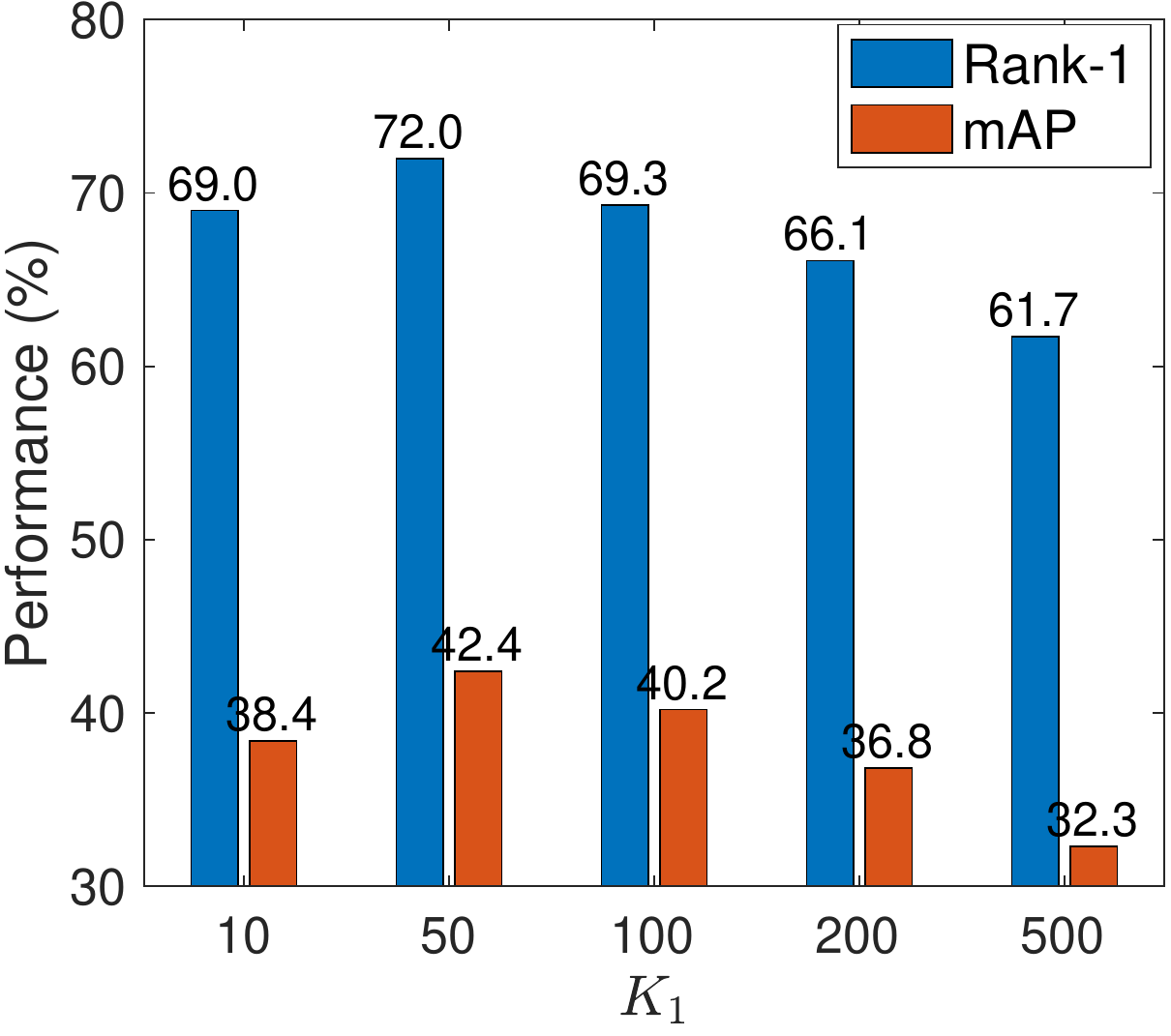}}
\caption{The performance evaluation with different values of $K_1$.}
\label{fig_analyze_k1}
\end{figure}

$K_1$ is the number of negative proxies mined in offline or online association. Figure~\ref{fig_analyze_k1} presents the performance of O2CAP when $K_1$ varies from 10 to 500. We observe that the performance goes up when the number of hard negatives increases from 10 to 50. But the performance gradually drops when more negative proxies are taken into consideration. It indicates that easy negatives may hamper the contrastive learning. Focusing on a small number of most informative negative proxies helps our model to better discriminate confusing instances. 

\subsubsection{The sensitivity of $K_2$} 
\begin{figure}[h]
\centering
\subfloat[\scriptsize{DukeMTMC-reID}]{\includegraphics[width=0.225\textwidth]{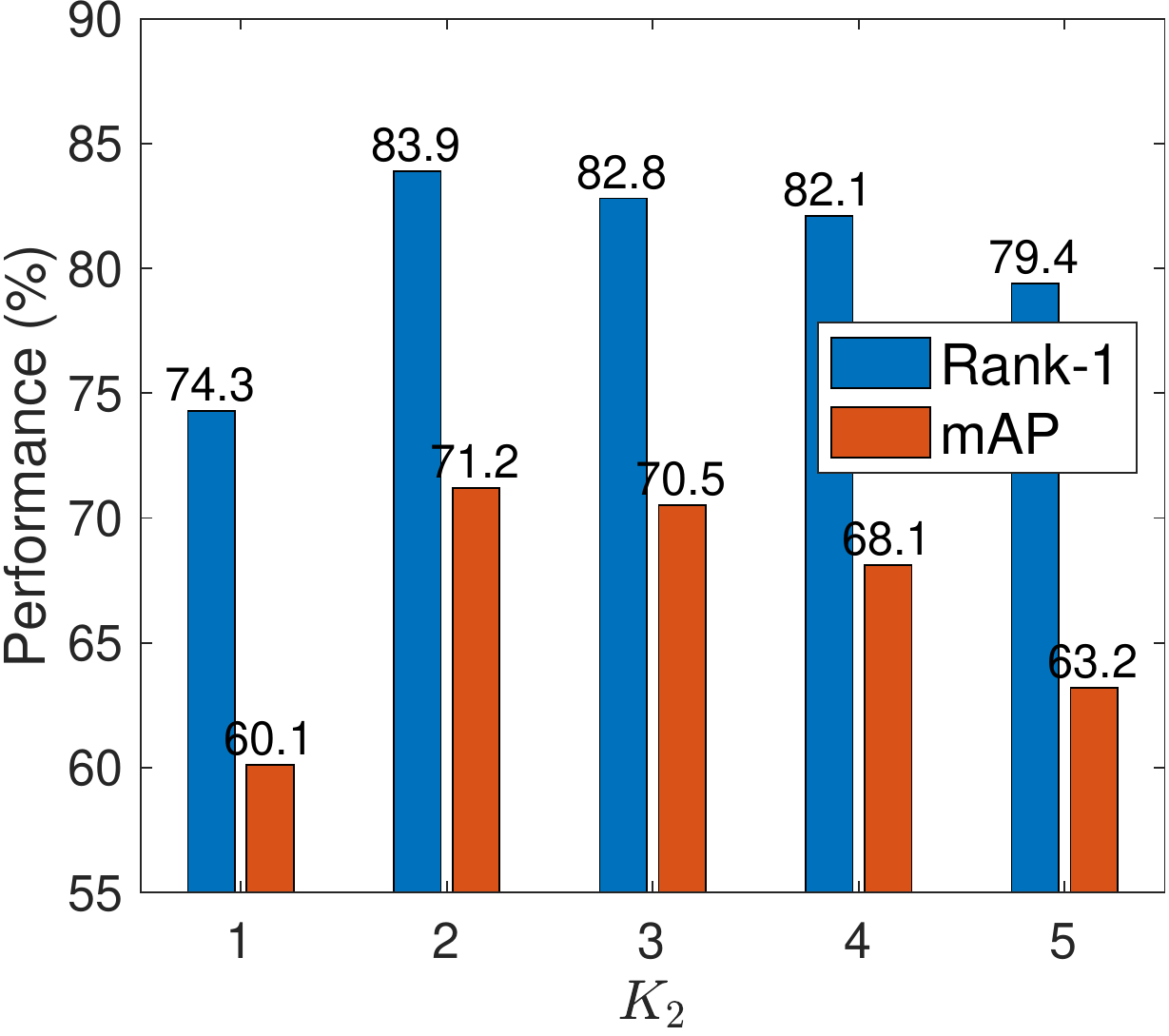}}
\quad
\subfloat[\scriptsize{MSMT17}]{\includegraphics[width=0.225\textwidth]{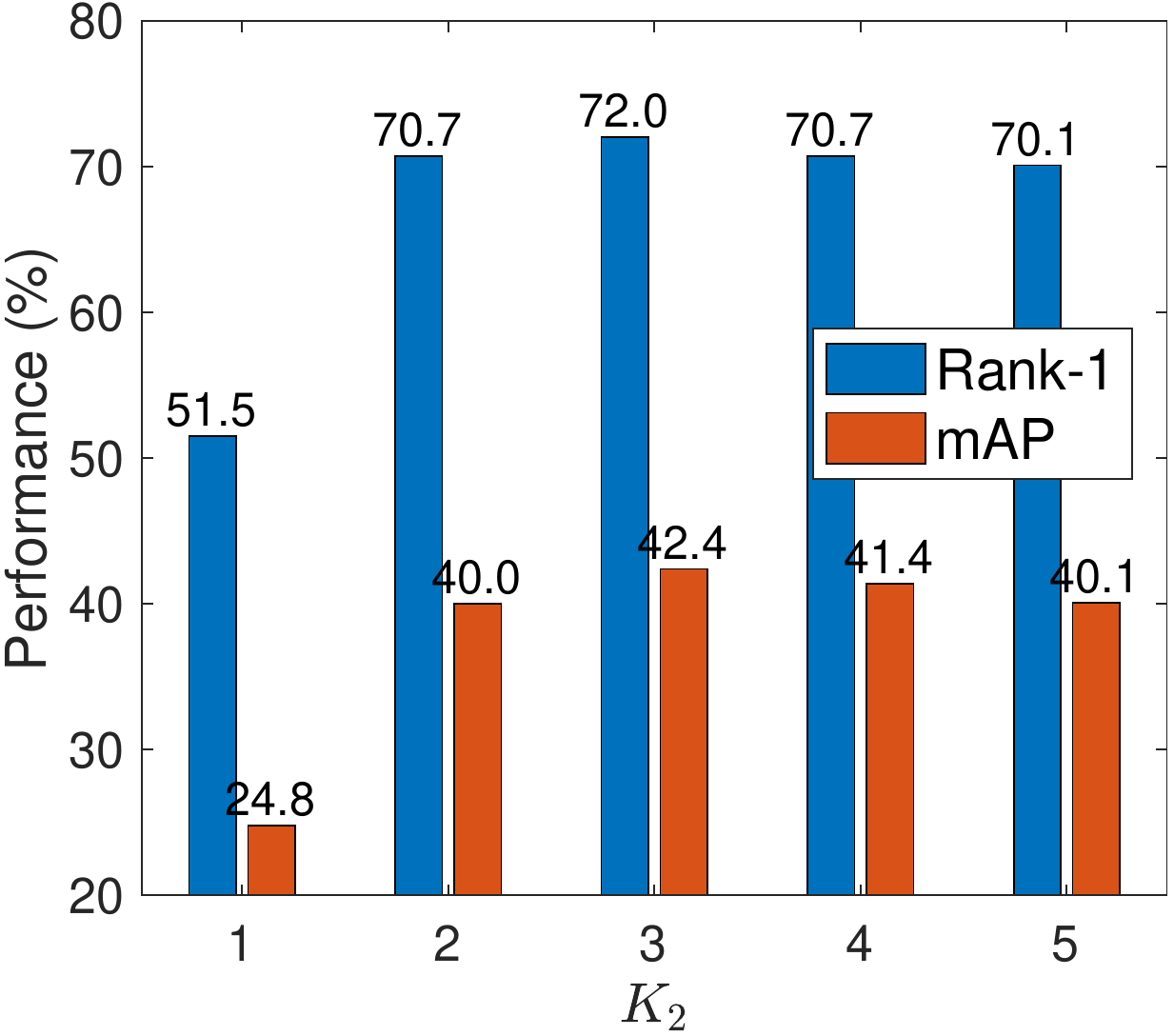}}
\caption{The performance evaluation with different values of $K_2$.}
\label{fig_analyze_k2}
\end{figure}

$K_2$ is the number of positive proxies associated online for each instance. Figure~\ref{fig_analyze_k2} presents the performance of O2CAP when $K_2$ varies from 1 to 5. We see that the performance increases first and then degenerates. We also notice that the best performed $K_2$ value is closely related to the CID value listed in Table~\ref{dataset_statistic_table}. If $K_2$ is larger than the CID value, false positives will be inevitably included and if $K_2$ is too small, no enough positive proxies will be recalled. Therefore, we set $K_2$ to be the value that is one less than CID for each dataset.  

\subsubsection{The sensitivity of $w$} 
\begin{figure}[h]
\centering
\subfloat[\scriptsize{DukeMTMC-reID}]{\includegraphics[width=0.225\textwidth]{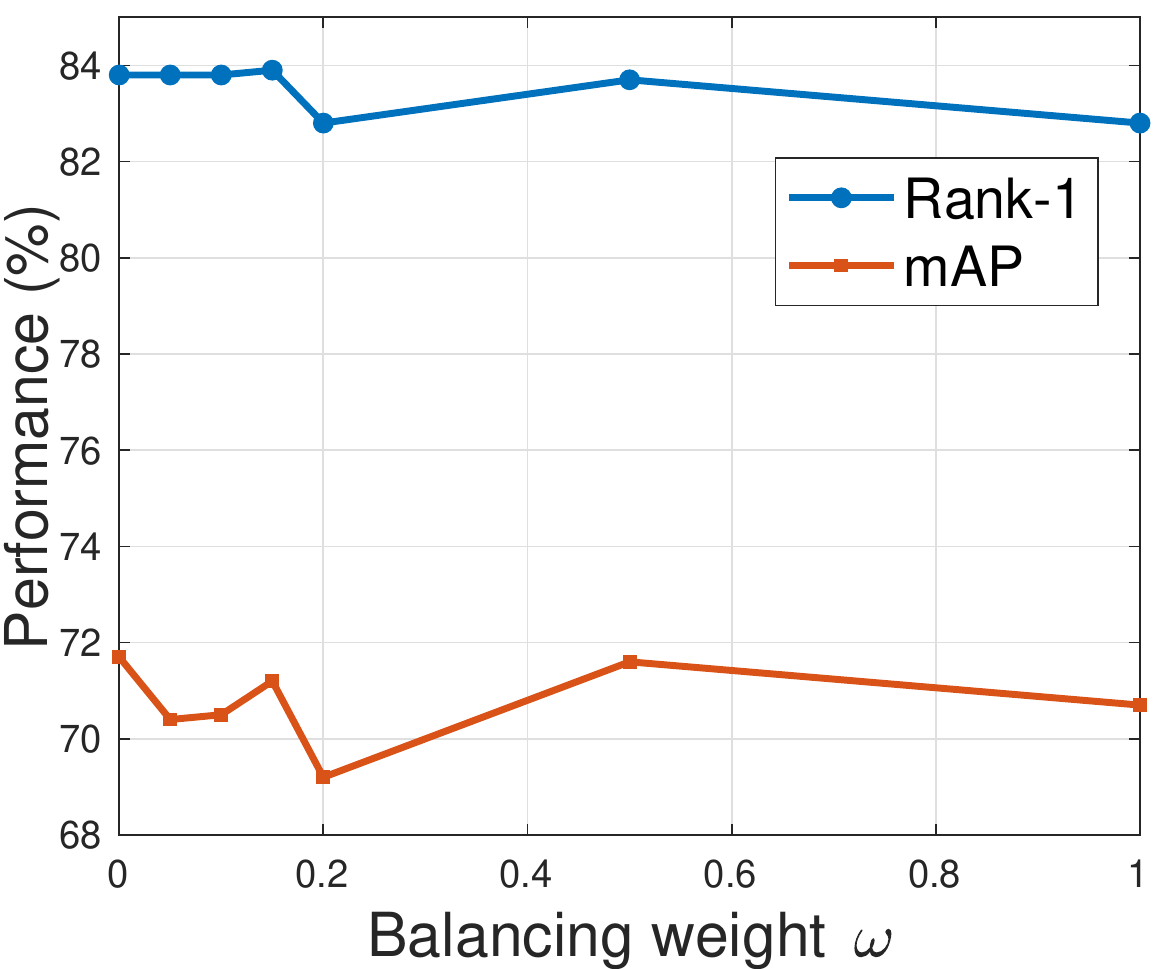}}
\quad
\subfloat[\scriptsize{MSMT17}]{\includegraphics[width=0.225\textwidth]{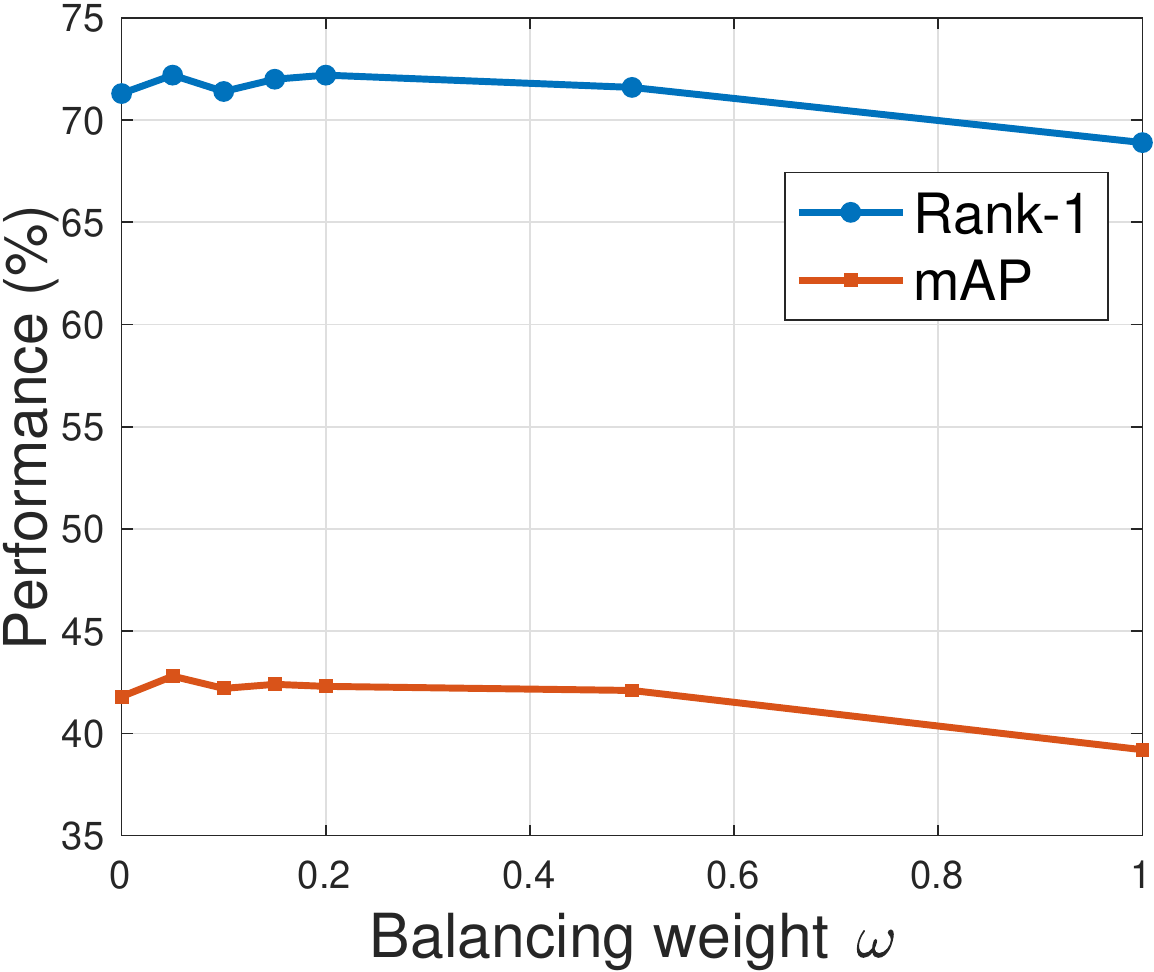}}
\caption{The performance evaluation with different values of $w$.}
\label{fig_analyze_w}
\end{figure}

The weight $w$ is to balance the instance-to-proxy and proxy-to-proxy parts in the balanced similarity. When $w = 1$, the balanced similarity degenerates to the instance-to-proxy similarity and when $w = 0$, it becomes the proxy-to-proxy similarity. Figure~\ref{fig_analyze_w} presents the performance varying along with the change of $w$. As shown in the figure, the performance on DukeMTMC-reID is less sensitive to $w$ when it increases from 0 to 1. On the contrary, setting $w$ to be a small value is more beneficial on MSMT17. Therefore, we set $w = 0.15$ throughout all other experiments to achieve a trade-off on all datasets.

\subsubsection{The sensitivity of training epochs} 
\begin{figure}[h]
\centering
\subfloat[\scriptsize{DukeMTMC-reID}]{\includegraphics[width=0.23\textwidth]{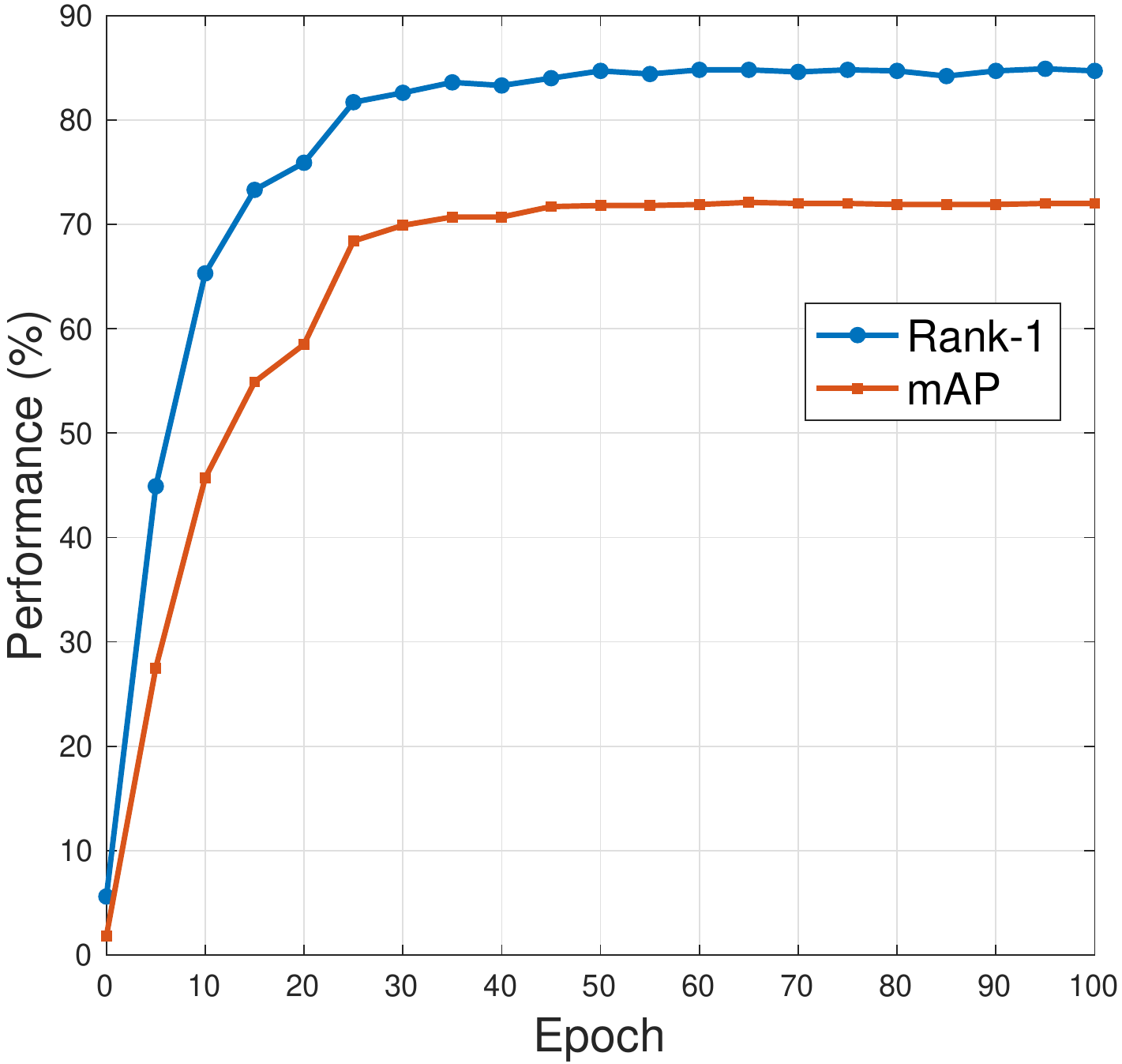}}
\quad
\subfloat[\scriptsize{MSMT17}]{\includegraphics[width=0.23\textwidth]{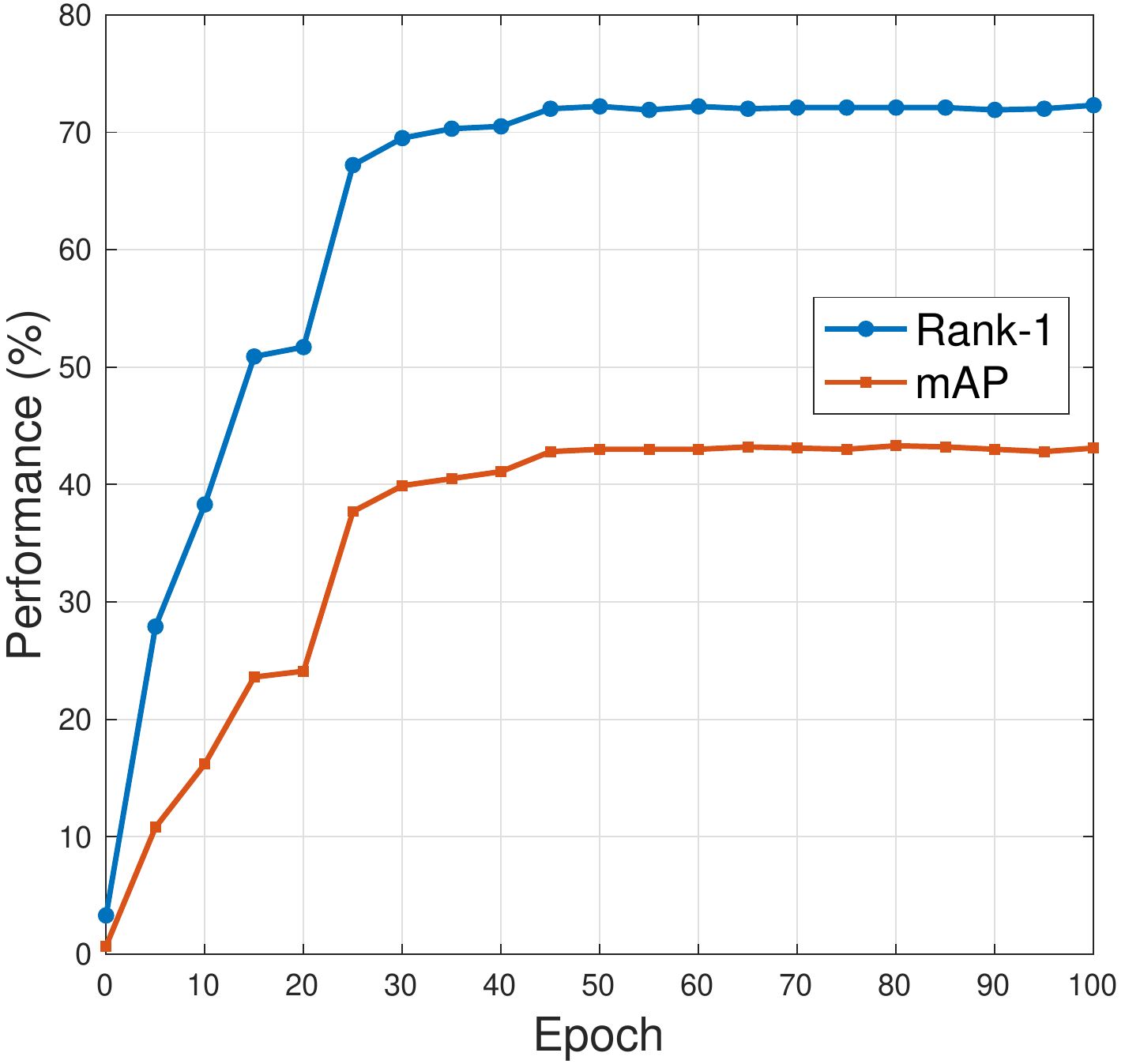}}
\caption{The performance evaluation with different training epochs.}
\label{fig_acc_epoch}
\end{figure}

Finally, we investigate the performance sensitivity with respect to the number of training epochs. Figure~\ref{fig_acc_epoch} presents the performance of O2CAP when the training epoch varies from 1 to 100. We observe that both Rank-1 and mAP go up quickly in the first 30 epochs, and then slowly converge at the 50-th epoch. After 50 epochs, the performance fluctuates very slightly and no discernible improvement is gained. Therefore, we set the total training epoch to 50 in order to achieve a trade-off between training time and model accuracy.

\begin{table*}[ht]
\centering
\caption{Comparison with state-of-the-art methods. Both purely unsupervised and UDA-based methods are included. We also provide several fully supervised methods for reference. O2CAP: our proposed method in this work. O2CAP(IBN): O2CAP with IBN-ResNet50 as backbone. O2CAP(IBN+GeMPool): O2CAP with IBN-ResNet50 and GeM Pooling. O2CAP(w/ GT): our proposed method with ground truth, indicating the upper bound.}
\scalebox{1.05}{
\begin{tabular}{c|c|cccc|cccc|cccc}
\hline  %\toprule
\multirow{2}{*}{Methods} & \multirow{2}{*}{Reference} &  \multicolumn{4}{c| }{Market-1501} & \multicolumn{4}{c|}{DukeMTMC-ReID}  & \multicolumn{4}{c}{MSMT17}  \\
\cline{3-14}&  & R1 & R5 & R10 & mAP   & R1 & R5 & R10  & mAP   & R1 & R5 & R10 & mAP \\ 
\hline
\multicolumn{13}{l}{\textbf{\textit{Purely Unsupervised}}} \\ \hline
BUC~\cite{lin2019aBottom}	& AAAI19        & 66.2 & 79.6 & 84.5  & 38.3              & 47.4 & 62.6 & 68.4 & 27.5         & - & - & - & - \\
UGA~\cite{wu2019graph} & ICCV19 	& 87.2 & - & - & 70.3 & 75.0 & - & - & 53.3 & 49.5 & - & - & 21.7 \\
SSL~\cite{lin2020unsupervised}	& CVPR20      & 71.7 & 83.8 & 87.4 &37.8	        & 52.5 &63.5 &68.9 & 28.6          & - & - & - & - \\
HCT~\cite{zeng2020hierarchical}    & CVPR20		& 80.0 & 91.6 & 95.2 & 56.4    & 69.6	& 83.4 & 87.4 & 50.7	         & - & - & - & - \\
CycAs~\cite{wang2020cycas} & ECCV20                  & 84.8 & - & - & 64.8 	                 & 77.9 &- & - & 60.1                    & 50.1 & - & - & 26.7 \\
IICS~\cite{Xuan2021cvpr} & CVPR21                  & 89.5  & 95.2 &  97.0 & 72.9 	 & 80.0  & 89.0  & 91.6  & 64.4          & 56.4 & 68.8 & 73.4  & 26.9   \\
RLCC~\cite{zhang2021refine}     & CVPR21     & 90.8 & 96.3 & 97.5 & 77.7           & 83.2 & 91.6 & 93.8 & 69.2    & 56.5 & 68.4 & 73.1 & 27.9 \\
\rowcolor{mygray} 
CAP~\cite{Wang2021CAP}        & AAAI21                & 91.4 &  96.3 &  97.7 &  79.2 &  81.1 &  89.3 &  91.8 &  67.3            &  67.4 &  78.0  &  81.4 &  36.9 \\ 
GroupSampling~\cite{groupsample21}  & arXiv21 & 92.3 & 96.6 &  97.8 & 79.2     & 82.7 & 91.1 &  93.5 & 69.1  & 56.2 & 67.3 & 71.5 & 24.6 \\
ICE~\cite{Chen2021ICE} & ICCV21 & 93.8 & 97.6 & 98.4 & 82.3 & 83.3 & 91.5 & 94.1 & 69.9 & 70.2 & 80.5 & 84.4 & 38.9 \\
ICE(IBN-ResNet)~\cite{Chen2021ICE} & ICCV21 & 94.2 & 97.6 & 98.5 & 82.5 & 83.6 & 91.9 & 93.9 & 70.7 & 70.7 & 81.0 & 84.6 & 40.6 \\
MGH~\cite{Wu2021MGH} & MM21 & 93.2 & 96.8 & 98.1 & 81.7 & 83.7 & 92.1 & 93.7 & 70.2 & 70.2 & 81.2 & 84.5 & 40.6 \\
MGCE-HCL~\cite{Sun2021} & EC21 & 92.1 &- & - & 79.6 & 82.5 & - & - & 67.5 & - & - & - & -\\
Liu \textit{et al.}~\cite{Liu2021} & arXiv21 & 93.0 & 97.5 & - & 82.4 & 84.9 & 92.3 & - & 72.2 & 68.6 & 79.4 & - & 38.4 \\ 
HHCL~\cite{9660560} & NIDC21 & 93.4 & 97.7 &  98.5 & 84.2    &  85.1 & 92.4 & 94.6 & 73.3   &  58.9 & 71.3 & 75.8 & 31.8 \\
\rowcolor{mygray}
O2CAP  & This work  & 92.5  & 96.9 & 98.0 & 82.7     & 83.9 & 91.3 & 93.4 & 71.2        & 72.0 & 81.9 &  85.4  &  42.4  \\
\rowcolor{mygray}
O2CAP(IBN)   & This work   & 93.1  & 97.4 & 98.1 & 83.7     & 85.2 & 91.9 & 93.5 & 72.8    & 75.5 & 84.8 & 87.7 & 46.9 \\
\rowcolor{mygray}
O2CAP(IBN+GeMPool)   & This work   & 93.3  & 96.9 & 97.7 & 85.0     & 85.3 & 91.4 & 93.3 & 73.2    & 77.3 & 85.6 & 88.2 & 48.3 \\
\hline 
\multicolumn{13}{l}{\textbf{\textit{Unsupervised Domain Adaptation}}} \\  \hline
PUL~\cite{unsup_clustering}  & TOMM18        & 45.5 &  60.7 &66.7 & 20.5         & 30.0 & 43.4 & 48.5 & 16.4           & - & - & - & - \\
SPGAN~\cite{Deng2018SPGAN}  & CVPR18    & 51.5 & 70.1 & 76.8 & 22.8       & 41.1 & 56.6 & 63.0 & 22.3          & - & - & - & - \\
ECN~\cite{zhong2019invariance}    & CVPR19      & 75.1 & 87.6 & 91.6 & 43.0      & 63.3 & 75.8 & 80.4 & 40.4    & 30.2 & 41.5 & 46.8 & 10.2\\
SSG~\cite{fu2019self}                    & ICCV19     & 80.0 & 90.0 & 92.4 & 58.3             & 73.0 & 80.6 & 83.2 & 53.4            & 32.2 & - & 51.2  & 13.3\\
AD-Cluster~\cite{zhai2020ad} & CVPR20 & 86.7 & 94.4 & 96.5 & 68.3 & 72.6 & 82.5 & 85.5 & 54.1 & - & - & - & - \\
MMCL~\cite{wang2020unsupervised} & CVPR20    & 84.4 & 92.8 & 95.0 & 60.4    & 72.4 & 82.9 & 85.0 & 51.4         & 43.6 & 54.3 & 58.9 & 16.2\\
Zhong \textit{et al.}~\cite{9018132} & TPAMI21 & 84.1  & 92.8 &  95.4 &  63.8     & 74.0 & 83.7  & 87.4  & 54.4      & 42.5 &  55.9 &  61.5 & 16.0\\
MMT~\cite{ge2020mutual}   & ICLR20              &87.7 & 94.9 & 96.9 & 71.2       & 78.0 & 88.8 & 92.5 & 65.1     & 50.1 & 63.9 & 69.8 & 23.3\\
MEB-Net~\cite{zhai2020multiple}   & ECCV20       & 89.9 & 96.0 & 97.5 & 76.0     & 79.6 & 88.3 & 92.2 & 66.1                & - & - & - & - \\
SpCL~\cite{ge2020self}        & NeurIPS20    &  90.3 &  96.2 &  97.7  &  76.7      &  82.9 &  90.1 &  92.5 & 68.8 &  53.1 & 65.8 &  70.5 & 26.5\\
SpCL(IBN-ResNet)~\cite{ge2020self} & NeurIPS20 & 91.5 & 96.9 & 98.0 & 79.2 & 83.4 & 91.0 & 93.1 & 69.9 & 58.9 & 70.4 & 75.2 & 31.8 \\
Isobe \textit{et al.}~\cite{Isobe2021} & ICCV21 & 94.2 & - & - & 83.4  & 83.5 & - & - & 70.8 & 66.6 & - & - &36.3  \\
Zheng \textit{et al.}~\cite{Zheng2021} & ICCV21 & 91.5 & - & - & 80.0 & 82.2 & - & - & 70.1 & 56.1 & - & - & 29.3 \\
MCRN~\cite{Wu2022} & AAAI22  & 93.8 & 97.5 &  98.5 & 83.8     & 84.5  & 91.7  & 93.8  & 71.5     & 67.5 & 77.9 & 81.6 & 35.7 \\
\hline
\multicolumn{13}{l}{\textbf{\textit{Fully Supervised}}} \\ \hline
PCB~\cite{sun2018beyond}   & ECCV18   & 93.8 &- &- & 81.6                  &83.3 &- &-   &69.2           &68.2 &- &- &40.4\\ 
ABD-Net~\cite{chen2019abd} & ICCV19 & 95.6 &- &- & 88.3         & 89.0 &- &- & 78.6          & 82.3 & 90.6 &- & 60.8 \\
FlipReID~\cite{Ni2021FlipReIDCT} & EUVIP21 & 95.3 & - & - & 88.5   & 89.4 & - & - & 79.8    & 83.3 & - & -& 64.3 \\
TransReID~\cite{He2021vit}  & ICCV21 & 95.2 & - & - & 89.5            & 90.7 & - & - & 82.6      & 86.2 & - & - & 69.4 \\
\rowcolor{mygray}
O2CAP(w/ GT)       & This work   & 94.4 & 98.0 & 98.8  & 86.1                 &88.2 & 94.1 & 96.0    &76.3         & 79.0 & 89.4 & 92.0  & 53.7 \\
\rowcolor{mygray}
O2CAP(IBN,w/ GT)  & This work   & 95.0 & 98.0 & 98.7  & 86.9                 &89.5 & 94.7 & 96.1   &78.3         & 81.0 & 90.3 & 92.7  & 56.7 \\
\rowcolor{mygray}
O2CAP(IBN+GeMPool,w/ GT)  & This work   & 95.0 & 98.0 & 98.6  & 87.3        &89.6 & 95.6 & 96.6   &79.9         & 82.8 & 91.1 & 93.0  & 58.8 \\
\hline
\end{tabular}
}
\label{compare_SOTA_table}
\end{table*}

\subsection{Comparison to State-of-The-Arts}
In this subsection, we compare the proposed method (named as O2CAP) with state-of-the-art methods on both person and vehicle Re-ID datasets. The comparison results are summarized in Table~\ref{compare_SOTA_table} and Table~\ref{compare_SOTA_table2}.

\subsubsection{Comparison with purely unsupervised methods on person Re-ID} 
14 representative or recent purely unsupervised methods are included for comparison. Most of these methods are clustering-based, and CAP~\cite{Wang2021CAP} (our preliminary work), ICE~\cite{Chen2021ICE}, MGH~\cite{Wu2021MGH}, MGCE-HCL~\cite{Sun2021}, Liu \textit{et al.}~\cite{Liu2021}, and HHCL~\cite{9660560} also adopt the contrastive learning technique same as this work. Both ICE and MGH are built upon CAP. The former extends CAP via introducing the instance-level contrastive learning and the latter additionally introduces hypergraph for label refinement. In contrast, our O2CAP sticks to the proxy-level contrastive learning and improves CAP via the combination of offline and online associations, keeping the model simple yet effective. From the results we see that O2CAP outperforms CAP and earlier methods by a significant margin. When compared to other contrastive learning based methods, O2CAP achieves competitive results on Market-1501 and DukeMTMC-ReID, while demonstrates a considerable superiority on MSMT17.

\subsubsection{Comparison with UDA-based methods on person Re-ID} 
Table~\ref{compare_SOTA_table} also presents 13 unsupervised domain adaptation based methods for comparison. Among them, ECN~\cite{zhong2019invariance}, SpCL~\cite{ge2020self}, Isobe \textit{et al.}~\cite{Isobe2021}, Zheng \textit{et al.}~\cite{Zheng2021} and MCRN~\cite{Wu2022} utilize contrastive learning as well. Although all UDA-based methods exploit external labeled data to boost the Re-ID performance on target datasets, they do not gain noticeable advantage when compared to recent purely unsupervised counterparts~\cite{Wang2021CAP,Chen2021ICE,Wu2021MGH}. For instance, our O2CAP performs on par or better than all UDA-based methods on DukeMTMC-ReID and MSMT17. Especially on the most complex dataset MSMT17, O2CAP outperforms them by a great margin due to the exploit of camera-aware proxies.

\subsubsection{Comparison with fully supervised methods on person Re-ID}
We additionally provide four representative fully supervised methods for reference, including PCB~\cite{sun2018beyond}, ABD-Net~\cite{chen2019abd}, FlipReID~\cite{Ni2021FlipReIDCT} and TransReID~\cite{He2021vit}. Besides, we also report the performance of our network backbone trained with ground-truth labels, which indicates the upper bound performance of our method. The results show that our unsupervised O2CAP has already outperforms the well-known PCB on all datasets with respect to all metrics, except Rank-1 on Market-1501. The performance gap between the unsupervised O2CAP and its supervised counterpart has also been greatly mitigated. Note that recent supervised methods such as FlipReID~\cite{Ni2021FlipReIDCT} and TransReID~\cite{He2021vit} have set up new state-of-the-art performance on all Re-ID datasets. It implies the potential of our unsupervised method to further improve performance if a more advanced backbone network could be adopted.

\subsubsection{O2CAP with a more advanced backbone}
The backbone used in O2CAP is ResNet50, which is a relatively plain network. In order to investigate the generalization ability of our method to different backbones, we additionally conduct an experiment using the IBN-ResNet50 backbone. It replaces batch normalization in ResNet50 via instance batch normalization (IBN)~\cite{Pan2018ibn}, which has been proved effective to boost the Re-ID performance~\cite{ge2020self,Chen2021ICE}. As shown in Table~\ref{compare_SOTA_table}, O2CAP with IBN-ResNet50 is able to improve the performance further. And using GeM (Generalized Mean) pooling together with IBN-ResNet50 gives additional improvement. Especially on MSMT17, 5.3\% Rank-1 and 5.9\% mAP improvements have been gained. The results show that our method is orthogonal to network design and could further benefit from better backbones.

\begin{table}[h]
\caption{Comparison with state-of-the-art methods on VeRi-776. $^\dagger$~indicates an UDA-based method working under the purely unsupervised setting.}
\label{compare_SOTA_table2}
\centering
\scalebox{0.9}{
\begin{tabular}{c|c|cccc}
\hline  %\toprule
\multirow{2}{*}{Methods} & \multirow{2}{*}{Reference} &  \multicolumn{4}{c}{VeRi-776}   \\
\cline{3-6} &  & R1 & R5 & R10 & mAP    \\ 
\hline
\multicolumn{6}{l}{\textbf{\textit{Purely Unsupervised}}} \\ \hline
SSML~\cite{Yu2021vehicle}  & IROS21 & 74.5 & 80.3 & - & 26.7 \\
SpCL$^\dagger$~\cite{ge2020self}          & NeurIPS20    & 79.9 & 86.8 &  89.9  & 36.9 \\
RLCC~\cite{zhang2021refine}     & CVPR21     & 83.4 & 88.8 & 90.9 & 39.6 \\
\rowcolor{mygray}
O2CAP  & This work  & 87.5&  92.7& 94.4& 41.9 \\
\rowcolor{mygray}
O2CAP(IBN)   & This work   & 89.6 & 93.5 & 94.7 & 42.4 \\
\rowcolor{mygray}
O2CAP(IBN+GeMPool)   & This work   & 89.7 & 93.8 & 95.1 & 43.0 \\
\hline
\multicolumn{6}{l}{\textbf{\textit{Fully Supervised}}} \\ \hline
VSCR~\cite{Teng2021viewpoint} & IJCV21 &  94.1 &  97.9 &  98.6 &  75.5 \\
VehicleNet~\cite{zheng2020vehicle} & TMM20     & 96.8 & - & - & 83.4 \\
TransReID~\cite{He2021vit}  & ICCV21 &             97.1 & - & - & 82.0 \\
HRCN~\cite{Zhao2021hetero}  & ICCV21 &             97.3 & 98.9 & - & 83.1 \\
\rowcolor{mygray}
O2CAP(w/ GT)     & This work            & 93.2  & 97.9  & 98.9  & 73.6 \\
\rowcolor{mygray}
O2CAP(IBN,w/ GT)  & This work   & 94.6 & 98.0 & 98.9 & 75.1 \\
\rowcolor{mygray}
O2CAP(IBN+GeMPool,w/ GT)  & This work   & 94.4 & 98.1 & 99.0 & 75.0 \\
\hline
\end{tabular}
}
\end{table}

\subsubsection{Comparison on vehicle Re-ID}
Table \ref{compare_SOTA_table2} presents the comparison results on a vehicle Re-ID dataset VeRi-776~\cite{liu2018veri}. Three state-of-the-art unsupervised methods including SSML~\cite{Yu2021vehicle}, SpCL$^\dagger$~\cite{ge2020self} and RLCC~\cite{zhang2021refine} are taken for comparison. Four fully supervised method VSCR~\cite{Teng2021viewpoint}, VehicleNet~\cite{zheng2020vehicle}, TransReID~\cite{He2021vit} and HRCN~\cite{Zhao2021hetero}, together with our network trained with ground truth, are also provided for reference. Compared to RLCC~\cite{zhang2021refine}, our method achieves 4.1\% Rank-1 and 2.3\% mAP improvements, validating its effectiveness for the unsupervised vehicle Re-ID task.

\section{Conclusion}
In this paper, we have presented a camera-aware proxy assisted method for the purely unsupervised person Re-ID. The proposed camera-aware proxies are able to deal with the large intra-ID variance via explicitly considering the variance within clusters resulted from the change of camera views. They can also better tackle the inter-ID similarity by paying more attention to the hardest negative instances when compared to the cluster-level counterparts. With the assistance of camera-aware proxies, two proxy-level contrastive learning losses based on offline and online associations are designed to optimize the Re-ID model. Extensive experiments on both person and vehicle Re-ID datasets, especially on the most challenging ones, have demonstrated the superiority of our method.

\section*{Acknowledgments}
This work was supported by Major Scientific Research Project of Zhejiang Lab, China (No. 2019DB0ZX01).

\bibliographystyle{IEEEtran}
\bibliography{reference_v3}

\end{document}